%% file: GSEM.tex
\pdfoutput=1
\documentclass[12pt]{article}

\newcommand{\thetitle}{Causal Modeling With Infinitely Many Variables}

\newcommand{\fullv}[1]{#1}
\newcommand{\shortv}[1]{}
\relax
\fullv{\usepackage[margin=1in]{geometry}}
\usepackage{times}
\fullv{\usepackage{graphicx}}
\graphicspath{ {.} }

\fullv{\usepackage{natbib}}
\renewcommand{\cite}{\citep}
\newcommand{\nciteyear}[1]{[\citeyear{#1}]}

\usepackage{times}

\usepackage{mathtools}
\usepackage{amssymb}
\usepackage{amsmath}

\shortv{\input{defn3}}
\fullv{\input{defn}}

\renewcommand{\S}{{\cal S}}
\renewcommand{\L}{{\cal L}}

\newcommand{\OFF}{\mathit{OFF}}
\newcommand{\ON}{\mathit{ON}}

\def\myvec{\mathaccent"017E}
\renewcommand{\YY}{\myvec{Y}}
\renewcommand{\yy}{\myvec{y}}
\renewcommand{\XX}{\myvec{X}}
\renewcommand{\xx}{\myvec{x}}
\renewcommand{\ZZ}{\myvec{Z}}
\renewcommand{\zz}{\myvec{z}}
\renewcommand{\WW}{\myvec{W}}
\renewcommand{\ww}{\myvec{w}}

\usepackage{bm}

\title{\thetitle}

\shortv{
\clearauthor{\Name{Spencer Peters} \Email{sp2473@cornell.edu}\and
 \Name{Joseph Y. Halpern} \Email{halpern@cs.cornell.edu}\\
 \addr Cornell University}
}
\fullv{
\author{Spencer Peters$^1$ \\ sp2473@cornell.edu \and Joseph Y. Halpern$^1$ \\ halpern@cs.cornell.edu}
\date{%
    $^1$Cornell University\\%
}
}

\begin{document}
\maketitle

\begin{abstract}


	Structural-equations models (SEMs) are perhaps the most commonly used
	framework
	for modeling causality. However, as we show, naively extending this framework to infinitely
	many variables, which is
	necessary, for example, to model dynamical
	systems, runs into several problems. We
introduce \emph{GSEMs} (\emph{generalized SEMs}),
a flexible generalization of SEMs
        that directly specify the results of interventions, in which
  (1) systems of differential equations can be represented in a
                natural and intuitive manner, (2) certain natural situations,
        which cannot be represented by SEMs at all, can be represented
        easily, (3) the definition of actual causality in SEMs carries
                over essentially without change.
\end{abstract}

\section{Introduction}
\label{sec:intro}

For
scientists trying to understand causal relationships, and policymakers
grappling with the consequences of their decisions, the structure of causality
itself is of great importance. One influential paradigm for formalizing
causality, \textit{structural-equations models} (SEMs),
describes
causal relationships as a collection of
\emph{structural equations.}
Actions taken by a scientist or policymaker
(interventions) manifest as modifications to the structural equations; for
example, if $X$ represents the rental price of an apartment, imposing rent
control amounts to replacing the equation for $X$ with $X = r$,
producing a new
SEM.

SEMs are well studied; there are standard
techniques for reasoning about
the outcomes of given interventions.
For example,
SEMs without cyclic dependencies
have a unique outcome for any given intervention,
which can be obtained simply by solving the equations in
any order consistent with the dependencies between
variables.
%
%
%
However, this property does not hold when there are infinitely many variables.
Consider the simple SEM with binary variables
$X_1, X_2, \dots$, where the
equations are $X_1 = X_2, X_2 = X_3, \dots$.%
\footnote{\label{footnote:extended-SEMs}SEMs are typically defined to
    have only finitely many variables; we relax this
	restriction
	for the purposes of illustration.}
Intuitively, this says that $X_1$ gets the value that $X_2$ has,
$X_2$ gets the value that $X_3$ has, and so on.
There are clearly no acyclic dependencies here; nevertheless,
these equations have
two solutions: $X_i = 0$
for all $i$, and $X_i = 1$ for all $i$.
In this case the solutions are easy to determine, but in more complex examples,
for example, a SEM with equations
$$X_{i} = \begin{cases}
 3 X_{i+1} + 1 & X_{i + 1} \text{ is odd} \\
 X_{i+1}/2 & X_{i + 1} \text{ is even,} \\
\end{cases}$$ it may be extremely difficult to determine the outcomes of interventions.
The problem here is that the dependency relation is
not well founded:
$X_1$ depends on $X_2$, which depends on $X_3$, and so on.
(A relation $\succ$ is well-founded if and only if there are no infinite descending sequences $X_1 \succ X_2 \succ X_3 \succ \dots$.)
The lack of well-foundedness is the source of the problem; it is easy to show
that if the dependency relation is
well-founded, then there will be a unique outcome.

Unfortunately, the problem of ill-founded dependencies is unavoidable
when working
with dynamical systems.
Suppose that we were to try to represent a dynamical system using a SEM.
Perhaps the most natural approach would
have variables $X_t$ representing the
state of a dynamical system at time $t$, where $t$ ranges over an
interval of real numbers. In general, we expect that changing the
value of $X_s$ will affect the value of $X_t$ for all $t > s$. That
is, the dependency relation can be identified with the less-than relation on
the real numbers, which is not well founded.

We
might try to avoid this problem by
choosing a finite timestep $\epsilon > 0$, and considering what
happens only at time steps that are multiples of $\epsilon$, using
structural equations of the form $X_{t + \epsilon} = f(X_t,
Y_t)$. This idea is reasonable and similar to practical
finite-difference methods for finding numerical solutions to dynamical
systems.  But it has several problems. Most importantly, the modeler
must design new approximate structural equations for each new
dynamical system and argue that these are
reasonable for the application at hand. The approach also inherits
issues from numerical computing that are best kept separate from
causal modeling issues to the extent possible.
These issues include the need to choose
$\epsilon$ (which depends on the particular system under study and the
questions of interest) and validate that the results are independent
of this choice.

%
To capture dynamical systems, we propose a more flexible class of
models that we call
\emph{generalized structural-equations models} (GSEMs).
It is easy to show that GSEMs indeed
generalize SEMs (see Theorem
\ref{thm:GSEM-generalizes-SEM}). Given a SEM and an
intervention, we can
produce a new SEM that represents the result of the intervention by
modifying the relevant equations.
GSEMs represent the same input-output relationships as SEMs---a GSEM
and an allowed intervention maps to a new GSEM, while a GSEM and a
context together determine a set of
possible outcomes (assignments to the endogenous
variables)---without committing to a specific mechanism for producing
the outcomes.
Indeed, any mapping from interventions and contexts to outcomes (all
defined with respect to some set of variables) is a GSEM. In this
sense, GSEMs are the most general causal model that has the same
``interface'' or input-output behavior as SEMs.
In fact, even if we restrict to settings with only finitely many variables,
GSEMs are more expressive than SEMs; see Example \ref{example:shell-game-SEM}.

Due to their generality, GSEMs no longer have some of the most
appealing features of SEMs. Most importantly, GSEMs
cannot be described using directed acyclic graphs, as SEMs can. Thus, many of
the graphical
tools developed for analyzing SEMs (such as the do-calculus)
do not apply to GSEMs.
However, we feel that this loss is worth accepting,
for two reasons. First, it is worth understanding what tools remain to
causal analysts when the graphical structure of SEMs is
removed. Second, and more significantly, we show that (unlike SEMs),
GSEMs can capture many standard formalisms for representing causality
in infinitary settings,
including
dynamical systems,
\emph{hybrid automata} \cite{alur_hybrid_1992}
(a popular formalism for describing mixed discrete-continuous systems),
and the \emph{rule-based models} commonly used
in molecular biology and organic chemistry
(see \cite{laurent_counterfactual_2018} and the references therein).
In particular, dynamical systems can be represented in a direct and
natural way using
GSEMs, avoiding all the problems mentioned above. The definition (see
Section \ref{sec:dyn-sys}) is
nothing more
than the textbook definition of the solution of a system of differential
equations.

\commentout{
In contrast, the most natural approach to modeling a dynamical system
with a SEM has several drawbacks. In this approach, one (roughly)
chooses a finite timestep $\epsilon > 0$ and designs approximate
structural equations that look like $X_{t + \epsilon} = f(X_t,
Y_t)$. This idea is reasonable and similar to practical
finite-difference methods for finding numerical solutions to dynamical
systems. But it has several problems. Most importantly, the modeler
must design new approximate structural equations for each new
dynamical system (rather than straightforwardly translating the exact
dynamics of the system into a GSEM) and argue that these are
reasonable for the application at hand. The approach also inherits
issues from numerical computing which are best kept separate from
causal modeling issues whenever possible (e.g., for the circuit ODE
discussed at the end of Section \ref{sec:dyn-sys} that is simple
enough to solve exactly). These issues include the need to choose
$\epsilon$ (which depends on the particular system under study and the
questions of interest) and validate that the results are independent
of this choice. Lastly, we note that this SEM, although technically
infinite, is functionally finite, in the sense that no matter what
(finite) set of variables the modeler is interested in, the infinite
``tail'' of the SEM extending past the last such variable is
irrelevant. This observation applies to all SEMs (due to
well-foundedness), so that the typical restriction of SEMs to finitely
many variables is essentially without loss of generality.
}

Thus, GSEMs serve as a unifying framework for many disparate
models of
causality.
Moreover, because GSEMs have the same interface as SEMs, any definition depending
only on inputs and outputs
(interventions and their outcomes)
can be immediately applied to GSEMs. In
particular, the notion of actual causality (whether event $X$ caused
event $Y$ in some concrete situation) given by Halpern and Pearl
\nciteyear{HP01b} and later modified by Halpern
\nciteyear{Hal47,Hal48} can be applied to GSEMs almost without
modification (see Section \ref{sec:ac}).
This means that causal modelers from, say, environmental scientists
studying predator-prey dynamical systems to molecular biologists
studying chemical reaction pathways, can use the same language of
definitions to describe actual causality and related notions.
These problems
can also be handled by \emph{causal constraints models}
(CCMs) \cite{BBM19}, which were introduced to get around
some of the restrictions of SEMs when modeling equilibrium solutions
of dynamical systems.  GSEMs and CCMs are actually equally expressive
(Theorem
\ref{theorem:CCMs-equivalent-to-GSEMs}).
However, the definition of GSEMs is  much simpler than the definition
of CCMs. Moreover, since GSEMs are a more straightforward
generalization of SEMs than CCMs, they are arguably easier to use for
those familiar with SEMs (as evidenced by the ease with which we carry
over the definition of actual causality).

\fullv{
The rest of the paper is organized as follows. In Section \ref{sec:sems}, we
review the standard SEM framework. In Section
\ref{sec:gsems}, we formally define GSEMs, relate them to
SEMs, and
describe some
of their advantages in more detail.
In Section \ref{sec:dyn-sys}, we show how to use GSEMs to model trajectories of
dynamical systems.
In Section \ref{sec:ccms}, we consider three other classes of models: causal
constraints models, hybrid automata, and rule-based models. We show
how GSEMs are a reformulation of causal constraints models, how they
can be used to answer causal questions about hybrid automata, and how
they complement existing \emph{ad hoc} causal modeling techniques defined for
rule-based models.
\shortv{Some material is deferred to
the full paper
  for reasons of space.}
}

\section{SEMs: a review}
\label{sec:sems}

Formally, a \emph{structural-equations model} $M$
is a pair $(\S,\F)$, where $\S$ is a \emph{signature}, which explicitly
lists the endogenous and exogenous variables  and characterizes
their possible values, and $\F$ defines a set of \emph{modifiable
	structural equations}, relating the values of the variables.
We extend the signature to include a set of \emph{allowed
	interventions}, as
was done in earlier work  \cite{BH19,Rub17}.

Intuitively, allowed interventions are the ones that are feasible or
meaningful.
A signature $\S$ is a tuple $(\U,\V,\R,\I)$.
$\U$ is a set of exogenous variables, $\V$ is a set
of endogenous variables, and $\R$ associates with every variable $Y \in
	\U \union \V$ a
nonempty, finite
set $\R(Y)$ of possible values for
$Y$ (i.e., the set of values over which $Y$ {\em ranges}). We
assume (as is typical for SEMs) that $\U$ and $\V$ are finite sets, and adopt the convention
that for $\YY \subseteq \U \cup \V$, $\R(\YY)$ denotes the product of
the ranges of the variables appearing in $\YY$; that is, $\R(\YY)
	\coloneqq \times_{Y \in \YY} \R(Y)$.
Finally, an intervention $I \in \I$ is a set of pairs $(X, x)$, where
$X \in \V$ and $x \in \R(X)$. For each $X \in \V$,
there is at most one $x \in \R(X)$ with $(X, x) \in I$.
We abbreviate an intervention $I$ by $\XX \gets \xx$, where
$\XX \subseteq \V$
and, unless $\XX$ is empty, $\xx \in \R(\XX)$.
Although this notation makes most sense
if $\XX$ is nonempty, we allow $\XX$ to be empty (which amounts to not
intervening at all).
If
$I$ consists of exactly one pair $(Y, y)$, we abbreviate $I$ as $Y
	\gets y$.

$\F$ associates with each endogenous variable $X \in \V$ a
function denoted $F_X$ such that $F_X: \R(\U \union \V - \{X\})
	\rightarrow \R(X)$.
This mathematical notation just makes precise the fact that
$F_X$ determines the value of $X$,
given the values of all the other variables in $\U \union \V$.
If there is one exogenous variable $U$ and three endogenous
variables, $X$, $Y$, and $Z$, then $F_X$ defines the values of $X$ in
terms of the values of $Y$, $Z$, and $U$.  For example, we might have
$F_X(u,y,z) = u+y$, which is usually written as
$X = U+Y$.   Thus, if $Y = 3$ and $U = 2$, then
$X=5$, regardless of how $Z$ is set.%

The structural equations define what happens in the presence of external
interventions.
Setting the value of some variable $X$ to $x$ in a SEM
$M = (\S,\F)$ results in a new SEM, denoted $M_{X
			\gets x}$, which is identical to $M$, except that the
equation for $X$ in $\F$ is replaced by $X = x$. Interventions on
subsets $\XX$ of $\V$ are defined similarly. Notice that $M_{\XX
			\gets \xx}$ is always well defined, even if $(\XX \gets \xx) \notin \I$.
In earlier work, the reason that the model included allowed
interventions was that, for example, relationships between two models
were required to hold only for allowed interventions (i.e., the
interventions that were meaningful).
As we shall see, here, the fact that we do not have to specify what
happens for certain interventions has a more significant impact.

Given a context $\uu \in \R(\U)$, the \emph{outcomes} of a
SEM $M$ under intervention $\XX \gets \xx$ are all
assignments of values $\vv \in \R(\V)$ such that the assignments $\uu$
and $\vv$ together satisfy the structural equations of
$M_{\XX \gets \xx}$.
This set of outcomes is denoted $M(\uu, \XX \gets \xx)$.
Given an outcome $\vv$,
we denote by $\vv[X]$ and $\vv[\XX]$ the value that $\vv$ assigns to $X$ and the restriction of $\vv$ to $\R(\XX)$ respectively.
We also use this notation for interventions; for example, $\yy[X]$ is
the value that intervention $\YY \gets \yy$ assigns to variable $X
	\in \YY$.

As discussed in the introduction, an important special case of SEMs
are acyclic (or recursive) SEMs.   Formally, an acyclic SEM is one
for which,
for every context $\uu \in \R(U)$,
%
there is some total ordering $\prec_\uu$ of the endogenous variables
(the ones in $\V$)
such that if $X \prec_\uu Y$, then $X$ is independent of $Y$,
that is,
$F_X(\uu, \ldots, y, \ldots) = F_X(\uu, \ldots, y', \ldots)$
for all $y, y' \in
\R(Y)$.  Intuitively, if a theory is acyclic, there is no feedback.
Acyclic models always have unique outcomes; this is a
consequence of assuming that $\V$ is finite.

In order to talk about SEMs and the information they represent more
precisely, we use the formal language $\L(\S)$ for SEMs
having signature $\S$, introduced by Halpern \nciteyear{Hal20}; see also
\cite{GallesPearl98}. An informal description of this language
follows; for more details, see \cite{Hal20}.
We restrict the language used by Halpern \nciteyear{Hal20} to formulas
that mention only allowed interventions.
Fix a signature $\S = (\U, \V, \R, \I)$.
Given an assignment $\vv \in \R(\V)$, the \emph{primitive event} $X = x$ is true of $\vv$, written $\vv \models (X = x)$, if $\vv[X] = x$; otherwise it is false. We extend this definition to \emph{events} $\phi$, which are Boolean combinations of primitive events, in the obvious way.
Given a SEM $M$ with signature $\S$ and an allowed intervention $\YY
	\gets \yy \in \I$, the \emph{atomic causal formula} $[\YY \gets
			\yy]\phi$ is true in context $\uu$, written $(M, \uu) \models [\YY
\gets \yy]\phi$ if, for all outcomes $\vv \in M(\uu,
        \YY \gets
	\yy)$, we have $\vv \models \phi$. Again, we extend this definition to
        \emph{causal formulas}, which are Boolean combinations of atomic
formulas, in the obvious way. The language $\L(\S)$ consists of all
causal formulas (over $\S$).
Using these formulas, we can also talk about properties that only
\emph{some} of the outcomes $\vv \in M(\uu, \YY \gets \yy)$
have.
 For an event
$\varphi$, define \mbox{$\< \YY \gets \yy \> \varphi$} as $\neg [\YY \gets \yy] (\neg
\varphi)$.
This formula is true exactly when $\varphi$ is true of at
least one outcome $\vv \in M(\uu, \YY \gets \yy)$.

The language of causal formulas completely characterizes the
outcomes of a causal model with finite outcome sets,
in the following precise sense.
(For the purposes of this paper, a \emph{causal model} is either a
SEM, a GSEM, or a
CCM.)

\begin{theorem}
	\label{theorem:satisfies-same-formulas-equiv-has-same-solutions}
	If $M$ and $M'$ are causal models over the same signature
	$\S$ that, given a context and intervention,
  return a finite set	of outcomes,
    then $M$ and $M'$ have the same outcomes (that
	is, for all $\uu \in \R(\U)$ and $I \in \I$, $M(\uu, I) =
		M'(\uu, I)$) if and only if they satisfy the same set of causal
	formulas (that is, for all $\uu \in \R(\U)$ and $\psi \in \L(\S)$,
	$M, \uu \models \psi \Leftrightarrow M', \uu \models \psi$).
\end{theorem}

\fullv{
The proof of this and all other results not in the main text can be
found in the
appendix.}
\shortv{The proof of this and all other results can be found in the full paper.}

\fullv{
	A short note on notation;
  consistent with the foregoing section, we use
	capital letters $X, Y$ to denote variables, lowercase letters $x, y$
	to denote the corresponding values;
  letters with arrows $\XX, \YY, \xx, \yy$ to denote vectors
  of variables and their corresponding vectors of
  values.
  We use boldface $\uu$ and $\vv$ for contexts and outcomes, respectively.
    We use $M, M'$ to denote causal models, script letters $\S, \U, \V, \R, \I$ to denote a model's signature and its components, script $\F$ for a SEM's structural equations, and boldface $\FF$ for a GSEM's outcomes mapping (see below).
}
\section{Generalized structural-equations models}
\label{sec:gsems}

The main purpose of causal modeling is to reason about a system's behavior
under intervention. A SEM can be viewed as a
function that
takes a context $\uu$ and an intervention $\YY \gets \yy$ and returns
a set
of
outcomes,
namely, the set of all solutions to the structural equations after replacing
the equations for the variables in $\YY$ with
$\YY = \yy$.

Viewed in this way, generalized structural-equations models (GSEMs) are a
generalization of SEMs. In a GSEM, there is
a function
$\FF$
that takes a context $\uu$ and an intervention $\YY \gets \yy$ and returns a
set of outcomes.
However, the outcomes need not be determined by solving a set of
suitably modified
equations as they are for SEMs. This relaxation gives GSEMs the
ability to concisely represent dynamical systems and other systems with infinitely many variables, and
the flexibility to handle situations involving finitely many variables that
cannot be modeled by SEMs.

Because GSEMs don't have the structure that SEMs have by virtue
of being defined in terms of structural equations, we may want to rule out certain unintuitive possibilities.
In particular, we require that after intervening to set
$\YY \gets \yy$, all outcomes satisfy $\YY = \yy$.
\subsection{GSEMs and SEMs}

Formally, a \emph{generalized structural-equations
  model (GSEM)} $M$ is a pair $(\S, \FF)$, where $\S$ is a signature,
and $\FF$ is a mapping from contexts and interventions to sets of
outcomes.
More precisely, a signature $\S$ is a quadruple $(\U, \V,
	\R, \I)$ where, as before, $\U$ is a set of exogenous variables, $\V$
is a set of endogenous variables, and $\R$ associates with every
variable $Y$ in $\U \cup \V$ a nonempty, finite set $\R(Y)$ of
possible values for $Y$; we extend $\R$ to subsets of $\V$ in the same
way as before. However, we no longer require that $\U$, $\V$ or the sets
$\R(Y)$ for $Y \in \U \cup\V$
be finite. The mapping $\FF$ is a function
$\FF: \I \times \R(\U) \to \P(\R(\V))$, where $\P$ denotes the powerset operation.
That is, it maps a context $\uu \in \R(U)$ and an allowed
intervention $I \in \I$ to a set of
\emph{outcomes} $\FF(\uu, I) \in \P(\R(\V))$.
As with SEMs, we denote these
outcomes
by $M(\uu, I)$.
As stated above, we require that outcomes
$\vv \in \FF(\uu, \XX \gets \xx)$ satisfy $\vv[\XX] = \xx$.
In
the special case where all interventions are allowed, we take $\I =
\I_{univ}$, the set of all interventions.
We note that, while these semantics are deterministic, we can bring
probability back into the picture just as we do for SEMs: by putting a
probability on contexts. To keep things simple, we consider only
deterministic examples in the
rest of the paper.

We now make precise the sense in which GSEMs generalize SEMs.
Two causal models $M$ and $M'$
are \emph{equivalent}, denoted $M\equiv M'$,
if they have the same signature and they have the same
outcomes,
that is, if for all sets of variables $\XX \subseteq \V$, all
values $\xx \in \R(\XX)$ such that $\XX \gets \xx \in \I$, and all contexts $\uu
	\in \R(\U)$, we have
$M(\uu, \XX \gets \xx) = M'(\uu, \XX \gets \xx)$.

\begin{theorem}\label{thm:GSEM-generalizes-SEM}
	For all SEMs $M$, there is a GSEM $M'$ such that $M \equiv M'$.
\end{theorem}

Just as for SEMs, the intervention $I = \YY \gets \yy$ on a GSEM $M$
induces another GSEM $M_I$.
To define $M_I$ precisely, we must first define the composition of
interventions.

Given interventions
$\XX \gets \xx$ and $\YY \gets \yy$, let their composition
$I = \XX \gets \xx; \YY \gets \yy$ be the intervention that
results by letting the intervention performed second ($\YY \gets \yy$)
override the first on variables that both
interventions
affect; that is,
$I = \XX \union \YY \gets \zz$, where for $Z \in \XX \union \YY$,
$$\zz[Z] =
	\begin{cases}
		\yy[Z] \quad \mbox{if $Z \in \YY$,} \\
		\xx[Z] \quad \mbox{if $Z \in \XX-\YY$.}
	\end{cases}$$
Given a GSEM $M = (\S, \FF, \I)$ and an intervention $I \in \I$,
define the intervened model $M_I$ to be
$(\S, \FF', \J)$, where
$\J = \{J \in \I_{univ} : I ; J \in \I\}$ and, for $J \in \J$,
$\FF'(\uu, J) = \FF(\uu, I ; J).$
(The same relationship holds between the signatures $\I$ of $M$ and
$\J$ of $M_I$ when $M$ is a SEM.)
%
Notice that if the set $\I$ is closed under composition, that is, if for all
$I, J \in \I$ we have $I ; J \in \I$, then $\J = \{J \in \I_{univ}
	: I ; J \in \I\} \supseteq \I$, so that with $M_I$ we have
all the interventions that we had with $M$, and perhaps more.

The skeptical reader may wonder if the mechanism of equation
modification in SEMs really is doing the same thing as the mechanism
of intervention composition in GSEMs. This is indeed the case.  There
are two equivalent ways to see this. The first is to show
that equation modification and intervention composition are the same
for SEMs.

\begin{theorem}
	\label{theorem:eqn-modification-equiv-intervention-composition-SEM}
	For all SEMs $M$ and interventions $I, J \in \I$ such that $I ; J
		\in \I$, we have that $M_I(\uu, J) = M(\uu, I ; J)$.
\end{theorem}

The second is to show that interventions respect equivalences that
hold between SEMs and GSEMs.
\begin{theorem}
	\label{theorem:intervened-models}
        If $M$ and $M'$ are causal models
  with $M \equiv M'$,
	then for all $I \in \I$, we have that $M_I \equiv M'_I$.
\end{theorem}
\subsection{Finite GSEMs}
\label{subsection:GSEMs-with-finitely-many-variables}
GSEMs clearly differ from SEMs in that the
sets of endogenous and exogenous
variables and the range of each individual variable can be
infinite. Consider the class of GSEMs where these restrictions are
retained, which we call \emph{finite GSEMs}. How do finite GSEMs
relate to SEMs? Halpern \nciteyear{Hal20} showed that all SEMs satisfy an
axiom system called $AX^+$%
\fullv{(see Appendix \ref{section:axioms}}
\shortv{(see the full paper)}
 for more details).
For example, one axiom (effectiveness)
states that after setting $X \gets x$,
all outcomes have $X = x$: $[\WW \gets \ww; X \gets x](X=x)$.
While we imposed this constraint explicitly on GSEMs (and hence this
axiom is \emph{valid} in GSEMs---it is true in all contexts of all
GSEMs), in SEMs there is
no need to impose it; it is a property of the way outcomes are calculated.
However, there are additional axioms, for example, one that requires
unique outcomes if we intervene on all but one endogenous variable,
that finite GSEMs do not satisfy. If we impose these
axioms on
finite
GSEMs, we recover SEMs.

\begin{theorem}
	\label{theorem:SEMs-are-equivalent-to-finite-GSEMs-if-all-interventions-allowed}
	For all finite GSEMs over a signature $\S$ such
	that $\I = \I_{univ}$ in which all the axioms of $AX^+$ are
        valid, there is an equivalent SEM,
	and vice versa.
\end{theorem}

Likewise, all acyclic SEMs satisfy an axiom system called $AX^+_{rec}$
\fullv{(also described in Appendix \ref{section:axioms}),}
\shortv{also described in the full paper}
which consists of the axioms in $AX^+$ along with two additional conditions.
Imposing these axioms on finite GSEMs when all interventions are
allowed recovers exactly the class of acyclic SEMs.

\begin{theorem}
	\label{theorem:GSEMs-and-acyclic-SEMs}
For all finite GSEMs over a signature $\S$
such that $\I = \I_{univ}$ and all the axioms $AX^+_{rec}$ are
valid, there is an equivalent acyclic SEM, and
vice versa.
\end{theorem}
We remark that the axiom system $AX^+$ can be
generalized so as to deal with arbitrary (not necessarily finite) GSEMs,
\fullv{and soundness and completeness results for GSEMs can be proved.
We defer these results to a companion paper \cite{PH20}.}
\shortv{and soundness and completeness results for GSEMs can be proved;
see \cite{PH20}.}


Theorems~\ref{theorem:SEMs-are-equivalent-to-finite-GSEMs-if-all-interventions-allowed} and
\ref{theorem:GSEMs-and-acyclic-SEMs}
show that finite GSEMs satisfying $AX^+$ and $AX^+_{rec}$, respectively,
are equivalent to SEMs and acyclic SEMs, respectively, if \emph{all interventions are allowed}.  This equivalence
breaks down once we restrict the set of interventions;
GSEMs are then strictly more expressive than SEMs, as the following
example shows.

\begin{example}\label{example:shell-game-SEM}
	Suppose that Suzy is playing a shell game with two shells. One of the
        	shells conceals a dollar; the other shell is empty.
        Suzy can choose to flip over a shell. If
	        she does, the house flips over the other shell. If Suzy picks shell 1,
    	which hides the dollar, she wins the dollar; otherwise she wins
	nothing. This example can be modeled by a GSEM $M_{shell}$ with
	two binary endogenous variables $S_1, S_2$
	describing whether shell 1 is flipped over and shell 2 is flipped
	over, respectively, and a binary endogenous variable $Z$ describing
	the change in Suzy's winnings.
	(The GSEM also has a trivial exogenous variable whose range has size
	1, so that there is only one context $\uu$.)
	That defines $\U, \V$, and $\R$;
	we set $\I = \{S_1 \gets 1, \S_2 \gets 1\}$; and $\FF$ is
	defined as follows:
	\begin{align*}
		\FF(\uu, S_1 \gets 1) & = M_{shell}(\uu, S_1 \gets 1)    \\
		                     & = \{(S_1 = 1, S_2 = 1, Z = 1)\}  \\
		\FF(\uu, S_2 \gets 1) & = M_{shell}(\uu, S_2 \gets 1)    \\
		                     & = \{(S_1 = 1, S_2 = 1, Z = 0)\}.
	\end{align*}
	$M_{shell}$ is clearly a valid GSEM. Furthermore, checking that
	$M_{shell}$ satisfies all the axioms in $AX^+$ is
	straightforward; see
  \fullv{
    Appendix \ref{appendix:additional-proofs} (Theorem \ref{theorem:shell-game-satisfies-AX})
    }
    \shortv{the full paper}
      for details.
  However, no SEM $M'$ with the same signature can have the
outcomes
  $M'(\uu, S_1 \gets 1) = \{(S_1 = 1, S_2 = 1, Z = 1)\}$ and
	$M'(\uu, S_1 \gets 2) = \{(S_1 = 1, S_2 = 1, Z = 0)\}$.
	This is because in a SEM,
	the value of
	$Z$ would be specified by a structural equation $Z = \F_Z(\U, S_1, S_2)$. This
cannot be the case here, since there are two outcomes having
	$S_1 = S_2 =
		1$, but with different values of $Z$.
	\bbox

\end{example}

This example shows that
finite GSEMs (even restricted to those satisfying the axioms of
$AX^+$) are more
expressive than SEMs
when not all interventions are allowed.
The fundamental issue here is that $Z$ is determined by the intervention (which shell Suzy picks), not the state of the shells. In SEMs, the system's behavior cannot depend explicitly on the intervention, only on the variables altered by the intervention.
We note that Suzy's situation can be modeled by a SEM with an
additional variable describing Suzy's action. (More precisely,
one where the only allowed interventions set this variable's value to match
Suzy's action.) However, this variable is redundant in the sense that
Suzy's action is already described by the intervention. Thus,
arguably, the GSEM model is more natural.



\shortv{\section{Dynamical systems}}
\fullv{\section{Ordinary differential equations}}
\label{sec:dyn-sys}

In this section, we show how GSEMs can be used to model dynamical
systems characterized by a system of ordinary differential equations
(ODEs).
Suppose that we have a system of ODEs of the form
\newcommand{\dvarv}[1]{\mathcal{#1}}
\newcommand{\ddvarv}[1]{\dot{\mathcal{#1}}}
\newcommand{\dvar}[1]{\mathcal{X}_#1}
\newcommand{\ddvar}[1]{\dot{\mathcal{X}}_#1}
\begin{align*}
	\ddvar{1} & = F_1(\dvar{1}, \dvar{2}, \dots, \dvar{n})  \\
	\ddvar{2} & = F_2(\dvar{1}, \dvar{2}, \dots, \dvar{n})  \\
	\dots     &                                             \\
	\ddvar{n}
	          & = F_2(\dvar{1}, \dvar{2}, \dots, \dvar{n}),
\end{align*}
where the $\dvar{i}$ are real-valued functions of time,
called \emph{dynamical variables},
and
$\ddvar{i}$ denotes the derivative of $\dvar{i}$ with respect to time.
%
(Nearly all systems of ODEs occurring in practice can be put into this
form by adding auxiliary variables \cite{YM2020}.
For example, $d^2\dvarv{X}/dt^2 =
	-\dvarv{X}$
becomes the pair of equations $d\dvarv{X}/dt = \dvarv{Y};
	d\dvarv{Y}/dt = -\dvarv{X}$.)
This system of ODEs, together with the initial values $\dvar{1}(0),
	\dvar{2}(0), \dots, \dvar{n}(0)$, determines a set of solutions over
the interval $[0, T]$ for
$T > 0$
or the interval $[0, \infty)$.
    We capture this system of ODEs using the GSEM
				\newcommand{\ODE}{\mathit{ODE}}
			$M_{\ODE} = (\S, \FF).$
				The signature $\S = (\U, \V, \R, \I)$ of $M_{\ODE}$ is defined as follows:

				%
			$\V = \{X_i^s: 1 \le i \le n, s \in
			(0,T]\}$, $\U = \{X_1^0, \ldots, X_n^0\}$,
$\R(V) = \mathbb{R}$ for $V \in \U \union \V$, and
$\I = \I_{univ}$.
Here the variable $X_i^s$ represents the value of the function
$\dvar{i}$ at time $s$, that is, $\dvar{i}(s)$.
The only nontrivial part of this
definition is the function $\FF$.
To describe $\FF$, we first need some preliminary definitions.
A dynamical variable $\dvar{i}$ is \emph{intervention-free with
	respect to $I = \XX \gets \xx$ on an interval $(a, b)$} if for all $t
	\in (a, b)$, we have $X_i^t
       	\notin \XX$; the interval $(a, b)$ is \emph{intervention-free} if all
        dynamical
variables are intervention-free on $(a, b)$.
Given a context $\uu$ and an
intervention $I = \XX \gets \xx$, let $\FF(\uu, I)$
consist of all outcomes $\vv$
that satisfy the following conditions.
Define the functions $\dvar{i}$ for $1 \leq i \leq n$ by taking
$\dvar{i}(t) = \vv[X_i^t]$.
We impose the following constraints:
\begin{description}
\item[ODE1.]
  The outcome $\vv$ agrees with $I$, that is, $\vv[\XX] = \xx$.
	\item[ODE2.]
	      For all $i$, $\dvar{i}$ is left-continuous except when intervened
	      on; that is, $\dvar{i}$ is left-continuous at all points $t$ such
	      that $X_i^t
		      \notin \XX$.

	\item[ODE3.] For all
	      intervention-free intervals $(a, b) \subseteq [0, T]$,
	      the functions $\dvar{1}(t), \dvar{2}(t), \dots, \dvar{n}(t)$ solve
	      the initial-value problem on $(a, b)$.
	      That is,
	      for all $1 \leq i \leq n$,
	      $\dvar{i}$ is
	      right-continuous at $a$,
	      differentiable
	      on $(a, b)$,
	      and its derivative $\ddvar{i}$ satisfies
	      $\ddvar{i}(t) = F_i(\dvar{1}(t), ..., \dvar{m}(t))$
	      for all $t \in (a, b)$.
\end{description}

These conditions require, rather straightforwardly, that the
outcomes agree with
the
differential equations except where the modeler has intervened
(and are appropriately continuous).
No fancy limits or partial functions are required.






This set of allowed interventions is rather rich.
In practical scenarios, we are typically interested in interventions where
\begin{itemize}
	\item a variable is set to a certain value at a certain
	      instant in time, or
	\item a variable is set to a certain value and held at
	      that value throughout an interval of time,
\end{itemize}
as well as finite compositions of these interventions.
%
Interventions on intervals interact particularly well with the
initial-value problems that arise in solving differential
equations. For these
interventions, it makes sense to demand that outcomes of the GSEM
satisfy a stronger version of condition ODE3 above.
Suppose that there are two dynamical variables $\dvarv{X}$ and
$\dvarv{Y}$. We have $\ddvarv{X} = \dvarv{Y}$ and $\ddvarv{Y} = \dvarv{X}$.
We are
interested in outcomes
on the interval $[0, 1]$. If we intervene to
set $X_t$ to 0 on the whole interval $[0, 1]$, applying ODE3, we would
find that any assignment whatsoever to the $Y_t$
can be extended to a valid outcome (along with $X_t = 0$). This is not
very useful
for modeling
purposes. It is more useful to require that the differential equation
for $Y$ still hold, and remove the differential equation for $X$.
This is the import of condition ODE3$'$ below.
A variable is
\emph{set to a constant $k$} during an open interval
$(a, b)$ if
for all $t \in (a, b)$,
$\xx[X_i^t] = k$. An open interval $(a, b)$ is
\emph{intervention-constant} if for all $1 \leq i \leq m$, $\dvar{i}$ is
either
intervention-free
on $(a, b)$ or set to a constant
$k$ (for some $k \in \mathbb{R}$)
during
$(a, b)$.
\begin{description}
  \label{description:ode3prime}
	\item[ODE3$'$.] For every intervention-constant open
	      interval $(a, b)$, if $\dvar{i}$ is intervention-free on $(a, b)$, then $\dvar{i}$
	      is right-continuous at $a$, differentiable on $(a, b)$, and
	      its derivative $\ddvar{i}$ satisfies
	      $\ddvar{i}(t) = F_i(\dvar{1}, ..., \dvar{m})(t)$ for all $t \in (a, b)$.
\end{description}
Notice that ODE3$'$ implies ODE3: if all
variables are
intervention-free on
$(a, b)$, then the outcomes
must satisfy all the differential equations on $(a, b)$.
We now show how to find the unique outcome in a GSEM
$M_{\ODE}$
satisfying ODE1,
ODE2,
and ODE3$'$
for a large class of interventions of practical interest that we
denote
$\I_{intervals}$.
$\I_{intervals}$ consists of all finite compositions $I_1; I_2; \dots;
I_m$, where each $I_j$ is either a point intervention $X_i^t \gets k$
%
or an interval intervention $ \{X_i^t \mid t \in (a, b)\} \gets k$,
which we abbreviate as
$X_i(a,b) \gets k$ for readability.
(We similarly
use the abbreviations $X_i[a, b)$, $X_i(a, b]$, and $X_i[a, b]$.)
Note that intervening on each of these sets can be achieved by
composing two or three
point or interval interventions.

Fix $I = I_1; \dots; I_m \in I_{intervals}$. Let $t_1 < t_2 < \dots <
	t_l$ be the endpoints of the intervals $I_1, \ldots, I_m$. (The
endpoint of $X_i^t \gets k$
is $t$ and the endpoints of $X_i(a, b) \gets k$ are $a$ and $b$.)
It is easy to see that each interval $(t_1, t_2)$ is intervention-constant.

%
%
Thus, we can find outcomes of the model step by step. The following
algorithm finds an outcome of $M_{\ODE}$ under intervention $I = \XX
\gets \xx \in \I_{intervals}$
with initial conditions $\uu = (X_1^0,
	\dots, X_n^0)$.
  Note that we take the ability to solve initial-value problems and
  store their solutions as primitive. For convenience, we define $t_0 = 0$.

  \begin{description}
	\item[Algorithm SOLVE-ODE-GSEM] \hfill
	      \begin{enumerate}
		      \item For $1 \leq i \leq n$, define $\dvar{i}(0) = X_i^0$.
		      \item For $i = 1, \dots, l$:
		            \begin{enumerate}
			            \item For $j = 1, \dots, n$, if $\dvar{j}$ is set to a constant $k$ on $(t_{i-1}, t_i)$,  define $\dvar{j}(t) = k$ for all $t \in (t_{i-1}, t_i)$.
			            \item Define the remaining (intervention-free) dynamical variables on $(t_{i-1}, t_i)$ so that
			                  for $j = 1, \dots, n$, if $\dvar{j}$ is intervention-free, then
			                  $\dvar{j}$ is right-continuous at $t_{i-1}$, differentiable on $(t_{i -1}, t_i)$, and its derivative $\ddvar{j}$ satisfies $\ddvar{j}(t) = F_j(\dvar{1}(t), \dots, \dvar{m}(t))$ for all $t \in (t_{i-1}, t_i)$.
  If there is no way to do this, output ``No solution''.
			                \item For $j = 1, \dots, n$, define $\dvar{j}(t_i)$ as follows.
\begin{enumerate}
\item[(i)] If $X_j^{t_i} \in \XX$, define $\dvar{j}(t_i) = \xx[X_j^{t_i}]$.
\item[(ii)] If $X_j^{t_i} \notin \XX$,
   define $\dvar{j}(t_i) = \lim_{t \to t_i^-} \dvar{j}(t)$.
			                  \end{enumerate}
		            \end{enumerate}
 \item Define the functions $\dvar{i}$, $1 \leq i \leq n$, on $(t_l, \infty)$ so that $\dvar{1}, \dvar{2}, \dots, \dvar{n}$ solve the initial-value problem on $(t_l, \infty)$,
		            as defined in ODE3.
		            Again, if there is no way to do this, output ``No solution''.
\item Output the outcome $\vv$
                defined by $\vv[X_i^t] = \dvar{i}(t)$ for all $1 \leq i \leq n$, $t > 0$.
	      \end{enumerate}
\end{description}
If all initial-value problems arising in steps 2(b) and 3 are uniquely
solvable, then SOLVE-ODE-GSEM outputs
the unique outcome $\vv \in M_{\ODE}(\uu, I)$. In the general
case where some initial-value problems have multiple (or no)
solutions,
SOLVE-ODE-GSEM is under-specified, because it may pick any valid
solution in steps 2(b) and 3. In this case, SOLVE-ODE-GSEM outputs all
the outcomes (and only the outcomes) $M_{\ODE}(\uu, I)$.
In particular, if the model has no outcome for intervention $I$
given initial conditions $\uu$, every execution of the algorithm
outputs ``No solution''.

\begin{theorem}
	\label{theorem:solve-ode-gsem}
  	The set of outcomes output by valid executions of
	SOLVE-ODE-GSEM are exactly the outcomes $M_{\ODE}(\uu, I)$.
\end{theorem}

An alternative approach to modeling this situation would allow ``partial
outcomes'': that is, whenever ``No solution'' would be output in
iteration $i$, instead define $\vv[X_j^t] = error$ for all $1 \leq j
\leq n$ and all $t > t_{i-1}$ (where $error$ is a special value
indicating that the differential equations to go have no solution).
The alternative model $M_{ODE}'$ with the outcomes
corresponding to this modified algorithm satisfies an analogue of
acyclicity, which we explore in a companion paper. (A very similar
reformulation yields a version of the hybrid automaton model in%
\fullv{Section \ref{sec:applications}}
\shortv{the full paper}
that also satisfies this acyclicity
condition.)

Note that while GSEMs satisfying ODE3 but not ODE3$'$ don't
in general
have
meaningful outcomes under interval interventions,
they do under point interventions; in fact, SOLVE-ODE-GSEM finds
the outcomes of such GSEMs under finite compositions of point
interventions.

We conclude this section by showing how a
textbook dynamical system---an LC circuit---can be modeled as a
GSEM. An LC circuit consists of a
voltage source, a capacitor, and an inductor.
The dynamical variable of interest is the charge $\dvarv{Q}(t)$ on the
capacitor; the voltage $V$, capacitance $C$, and inductance $L$
are fixed parameters (although we encode them as dynamical variables
with zero derivatives).
The differential equations governing this circuit's behavior are
\begin{align*}
  \ddvarv{Q}(t) &= \dvarv{K}(t) \\
  \ddvarv{K}(t) &= \frac{\dvarv{V}}{\dvarv{L}} - \frac{1}{\dvarv{LC}} \dvarv{Q}(t) \\
  \ddvarv{V}(t) &= \ddvarv{C}(t) = \ddvarv{L}(t) = 0,
\end{align*}
where $K$ is the current.
 The solutions of these differential equations (for $\dvarv{Q}(t)$) take the form $$\dvarv{Q}(t) = \dvarv{V} \dvarv{C} + A \cos(\omega t + B),$$
where $\omega = 1/\sqrt{\dvarv{L}\dvarv{C}}$, and $A$ and $B$
are determined by the initial conditions on $\dvarv{Q}$ and
$\dvarv{K}$: $B = \arctan(\frac{\dvarv{K}(0)}{\omega(\dvarv{V} \dvarv{C} -
  \dvarv{Q}(0))})$ and $A = \frac{\dvarv{Q}(0) - \dvarv{V}
  \dvarv{C}}{\cos(B)}$. Note that these expressions make sense for
all initial conditions except when $\dvarv{V} \dvarv{C} -
\dvarv{Q}(0) = 0$; in this case the solution is instead
$\dvarv{Q}(t) = \dvarv{V} \dvarv{C}$ for all
$t$.

\begin{figure}
    \centering
   \shortv{\includegraphics[width=0.4\textwidth]{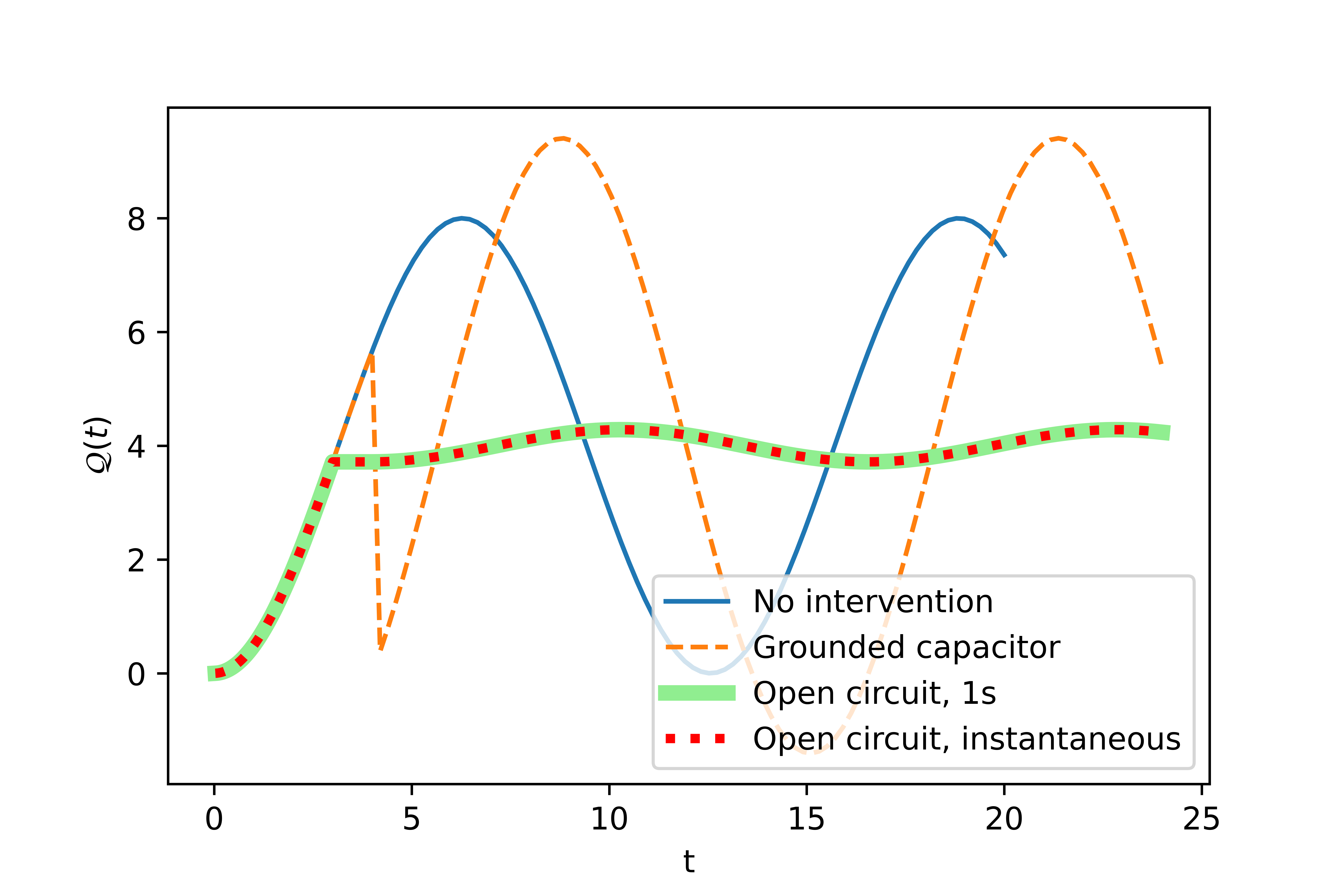}}
   \fullv{\includegraphics[width=0.8\textwidth]{solutions-plot-for-bw}}
    \caption{LC circuit outcomes under different interventions.}
        \label{figure:solutions}
\end{figure}

It follows from the explicit forms of
  the solutions given above that all initial-value problems involving
  these differential equations have unique solutions. It is similarly
  easy to see that if any of the differential equations (for variable
  $\dvarv{X}$)  is replaced with $\ddvarv{X} = 0$, all initial-value problems involving the resulting modified system of differential
  equations also have unique solutions.
  Thus, for all contexts $\uu$ and $I \in \I_{intervals}$,
  SOLVE-ODE-GSEM returns the unique
  outcome $\vv \in M(\uu, I)$.
    Suppose that the initial conditions of
  the circuit are given by the context
$\uu = \{Q^0 = 0, K^0 = 0, V^0 = 2, C^0 = 2, L^0 = 2\}$.
The unique outcome $\vv \in M(\uu, \emptyset)$ under the empty intervention is shown in Figure \ref{figure:solutions} (blue curve).

  Now suppose that the capacitor breaks down if $\dvarv{Q}$ exceeds
  6. Figure 1 shows that in the absence of intervention, $\dvarv{Q}$
  exceeds 6 just after $t = 4$. To prevent this, the operators ground
  the capacitor at time 4 (i.e., make the intervention $Q^4 \gets
  0$). However, this does not help. As shown in Figure
  \ref{figure:solutions} (orange curve), this initially reduces
  $\dvarv{Q}$, but eventually it exceeds 6. Next, the operators try
  opening the circuit at time 3 and closing it again at time 4%
  ; that is, performing the intervention $K(3, 4) \gets 0$. (Recall that $K(3, 4) := \{K^t \mid t \in (3, 4)\}$.)
  This results in the circuit
  entering an operating regime where $\dvarv{Q}$ is nearly constant
  (the green curve in Figure \ref{figure:solutions}), and never
  exceeds 6. We can show that, in agreement with intuition, the first
  intervention is not a cause of the capacitor not breaking down, and
  neither is the second
(because it is not the minimal intervention required to bring about
    the outcome).
  But a sub-intervention
  $K^4 \gets 0$ of the second intervention is
  a cause
  (corresponding to the red curve in Figure \ref{figure:solutions}).
  See Section \ref{sec:ac} for details.
\section{Related modeling techniques}
\label{sec:ccms}

We designed GSEMs as an extension of SEMs that can model
systems with a continuous-time component. Many other causal or
mechanistic modeling schemes have been proposed for such systems in
the literature. In this section, we review three of these schemes:
causal constraints models \cite{BBM19}, hybrid automata \cite{alur_hybrid_1992}, and counterfactual traces in rule-based models \cite{laurent_counterfactual_2018},
and discuss their relationship to GSEMs.
%
\shortv{
  (For space reasons, we defer the discussion of
  hybrid automata and rule-based models to the
full paper.)
} 
We remark that Ibeling and Icard \nciteyear{II19} consider SEMs
with infinitely many variables and a second class of models that they
call \emph{monotone simulation models}.  They show that monotone
simulation models are equivalent to \emph{computable} SEMs (ones
where, roughly speaking, the structural equations are computable
functions).  Our examples showing that GSEMs are more general than
SEMs apply immediately to the models considered by Icard and Ibeling.

\subsection{Causal constraints models}
At a coarse-grained level of modeling, such as when describing
equilibrium solutions of dynamical systems, constraints between
variables are natural objects of study. In order to describe
equilibrium solutions and functional laws (roughly, dependencies that
cannot be violated via intervention), \cite{BBM19} introduced the
notion of a causal constraints model (CCM). These models are composed of a
set of constraints, each of which are active only under selected
intervention targets; their outcomes
under a given intervention are the solutions of the constraints active
under the intervention's target.

A CCM
$M$ can be viewed as a pair $(\S,\C)$, where $\S = (\U, \V, \R, \I)$ is a signature
and $\C$ is a set of causal constraints.%
\footnote{
  The definition of CCMs in \cite{BBM19} does not include a set of
  allowed interventions. We include one here (in effect, generalizing
  CCMs slightly) to allow for a fair comparison with GSEMs.
}
Each
constraint $C \in \C$ is a pair $(f_C, a_C)$, where
$f_C:\R(\U) \times \R(\V) \to \{0, 1\}$ and $a_C \subseteq \P(\V)$.
Given a context $\uu$ and an
intervention $\XX \gets \xx \in \I$, the
outcomes $M(\uu, \XX \gets \xx)$ are all assignments $\vv$ to the
variables of $\V$ such that, for all $C
\in \C$ with $\XX \in a_C$,
$C$ is satisfied (e.g., we have  $f_C(\uu, \vv) = 1$) and
$\vv[\XX] = \xx$.
\footnote{
  The semantics given here is equivalent to the semantics in
  \cite{BBM19}, but we simplified the exposition slightly. In
  particular, in \cite{BBM19}, $f_C$ is allowed to map to an arbitrary
  measurable space, each constraint has an additional constant $c_C$,
  and $C$ is satisfied if $f_C(\uu, \vv) = c_C$. In addition, rather
  than just requiring that $\vv[\XX] = \xx$, Blom et al.~add
  constraints to enforce this condition.
  The changes that we have made do not affect the expressive power of CCMs.
We do not give the semantics of the intervened CCMs $M_I$, since they can be
derived from the semantics of outcomes in exactly the same way that we
  derived the semantics of intervened GSEMs in Section \ref{sec:gsems}
  (using composition of interventions).
}
Causal constraints models are equivalent to GSEMs.
\begin{theorem}
	\label{theorem:CCMs-equivalent-to-GSEMs}
	For all GSEMs, there is an equivalent
	CCM, and vice versa.
\end{theorem}
Of course, if we restrict to CCMs that satisfy the axioms in $AX^+$
(resp. $AX^+_{rec}$), we can prove an analogue of
Theorem~\ref{theorem:CCMs-equivalent-to-GSEMs}
for GSEMs satisfying $AX^+$ (resp. $AX^+_{rec}$).

CCMs were designed for characterizing equilibrium solutions of
dynamical systems.  They seem
less well suited for our intended applications, since they do not simplify
the process of reasoning from the model specification (i.e., constraints) to
solutions.
Thus, even though they are equivalent in expressive power, we believe
that CCMs and GSEMs will find complementary applications.

\section{Actual causes}
\label{sec:ac}

One important application of causal modeling is to deducing the
\emph{actual cause(s)} of $X = x$, that is (roughly speaking) the specific reasons that $X$
takes value $x$ in a given context $\uu$ and outcome $\vv$.

Many definitions of actual cause in SEMs have been proposed (e.g.,
\cite{BV18,GW07,hitchcock:99,Hitchcock07,Weslake11,Woodward03}). For
definiteness, we use that of Halpern and Pearl \nciteyear{HP01b}, as
later modified by Halpern \nciteyear{Hal48}, except that we further modify it
to deal with allowed interventions (which were not considered in
earlier definitions).

\begin{definition}[Actual cause]
  \label{definition:ac}
	%
	%
	Given a causal model $M$, $\uu \in \R(\U)$, and $\vv \in \R(\V)$,
	$\XX = \xx$ is an \emph{actual cause} of
	the event
	$\varphi$
	in $(M, \uu, \vv)$
	if the following three conditions hold:
  \begin{description}
\item[AC1.]
    $\vv \models \XX = \xx$ and $\vv \models \varphi$.
\item[AC2.]
    There is some $\WW \subseteq \V$ and a setting $\xx'$
		      of $\XX$ such that
		      $\WW \gets \vv[\WW] ; \XX  \gets \xx' \in \I$ and
		      $(M, \uu) \models \<\WW \gets \vv[\WW] ; \XX \gets \xx'\> \neg \varphi.$
        \item[AC3.]
          No proper subset of $\XX$ satisfies conditions AC1 and AC2.
\end{description}
\end{definition}


Intuitively, this definition captures the fact that when reasoning
about counterfactuals in a concrete scenario, we often want to fix
some details $\WW$ to the values $\vv[\WW]$ that they actually had in
that scenario; see \cite{Hal47} for more examples
and motivation. That paper did not consider allowed interventions.
To deal with allowed interventions, we insist that the intervention
$\WW \gets \vv[\WW] ; \XX
	\gets \xx'$ appearing in AC2 above is allowed.

  Using this formal definition of causality, we can verify the claims made in
  Section \ref{sec:dyn-sys} that (1) the intervention $Q^4 \gets 0$
  is not an (actual) cause
    of the capacitor not breaking down, but (2) the intervention $K^4
  \gets 0$ is.
  %
Recall that the capacitor breaks down if $\dvarv{Q}$ exceeds 6, so the
capacitor not breaking down corresponds to the statement
$\phi = \forall t > 4, \; \dvarv{Q}(t) \leq 6.$%
\footnotemark\footnotetext{
  \label{footnote:ac}
  This statement $\phi$, since it is universally quantified over all
  $t > 4$,  cannot be expressed in the language of causal formulas
  $\L(\S)$ we defined in Section \ref{sec:sems}. However, we do not
  view this as a serious problem. The definition of actual causality
  (Definition \ref{definition:ac}) allows arbitrary $\phi$, and it
  still makes sense if we take $\phi$ to be a formula in a richer
  language containing the first-order quantification we need here.
}
  Claim 1 is obvious, because under the first
  intervention, $Q^t$ does exceed 6 (at, say, $t=10$).
  Thus $\vv \models \neg \phi$,
    where $\vv$ is the outcome under $Q^t \gets 0$, which violates AC1.
    For Claim 2, note that
  on intervention-free intervals,
  the solutions are periodic in time, with period $2 \pi / \omega = 2
      \pi \sqrt{L^0 C^0} = 4 \pi \approx 12.6$. Since
  $\dvarv{Q}$ does not exceed 6 on the interval $(4, 4 + 12.6)$, it
  will never exceed 6.
Thus $\vv \models \phi$,
where $\vv$ is the outcome $K^4 \gets 0$,
  satisfying AC1. AC2 is
  satisfied, because if we instead set $K^t \gets K(t)$ for $t \in (3,
  4)$  where $K(t)$ is the value $K^t$ had under the empty
  intervention, $Q^t$ does exceed 6 (at, say, $t = 6$). (Here we are
  taking $\WW = \emptyset$.)
  Finally, AC3 is satisfied, because the intervention $K^4 \gets 0$ is on a single variable, and therefore minimal.
  As mentioned before, we deferred several more examples applying this definition of actual cause in dynamical systems (hybrid automata and rule-based models) for reasons of space.

    \newcommand{\init}{\mathsf{init}}
  \newcommand{\jump}{\mathsf{jump}}
  \newcommand{\flow}{\mathsf{flow}}
  \newcommand{\inv}{\mathsf{inv}}
  \newcommand{\dvarvb}[1]{\bm{\mathcal{#1}}}
  \newcommand{\ddvarvb}[1]{\bm{\dot{\mathcal{#1}}}}
  \renewcommand{\xi}{\dvarv{X}_i}
  \newcommand{\dxi}{\ddvarv{X}_i}
  \newcommand{\xs}{\dvarvb{X}}
  \newcommand{\dxs}{\ddvarvb{X}}
  \newcommand{\temp}{\dvarv{T}}
  \newcommand{\dtemp}{\ddvarv{T}}
  \newcommand{\dd}{\dvarv{D}}
\newcommand{\bound}{\mathsf{Bound}}
\newcommand{\phos}{\mathsf{Phos}}
\newcommand{\targ}{\mathsf{Targ}}
\newcommand{\trace}{\mathsf{Trace}}

\newcommand{\sectionDescribingOtherDynamicalSystems}{
\section{Other dynamical systems models}
  \label{sec:applications}
  \subsection{Hybrid automata}


Hybrid automata are a well-developed class of models for systems that have both continuous and discrete components
\cite{alur_hybrid_1992}; for example, a thermostat
controlling a heater to keep the temperature within a certain
tolerance of a set point. The state variables for this system are both
continuously varying in time (the temperature) and discretely changing
in time (whether the heater is on).
%
In this section, we show how to construct a GSEM model corresponding
to an arbitrary hybrid automaton. We demonstrate our construction on a
simple example from \cite{henzinger_theory_2000} and show how the
resulting GSEM can be used to answer causal questions.

Mathematically, a hybrid automaton is a finite directed multigraph $G
= (V, E)$,
a set $\dvarvb{X} = \{\dvarv{X}_1, ..., \dvarv{X}_n\}$ of real-valued
\emph{dynamical variables},%
\footnote{
  In the literature, these are usually just called variables; we call them dynamical variables to avoid confusing them with GSEM variables.}
and some predicates (discussed
below).%
\footnote{There is also a set of events $\Sigma$ used to
	disambiguate discrete transitions, but this is not relevant for us.}
The states $v \in V$ are called \emph{control modes}, and they represent the state of the discrete component of the system.
The edges $e = (u, v, n) \in E$,
where $u$ and $v$ are control modes and $n$ indexes the edge among all edges from $u$ to $v$,
are called \emph{control switches}, and they describe transitions between control modes.
Recall that a multigraph is a graph which may have multiple edges between any given pair of nodes; different control switches between the same pair of control modes represent different modes of transition between them. For example, the heater may have multiple triggers that change its state from OFF to ON.

The semantics of a hybrid automaton is defined by
predicates $\init, \flow, \jump$, and $\inv$. Possible starting configurations are given by
$\init(v, \dvarvb{X})$. Possible continuous dynamics within a control
mode are given by $\flow(v, \xs, \dxs)$, where $\dxs$
represents the vector of first derivatives. Possibly discontinuous
changes are given by $\jump(e, \xs, \xs')$, where $\xs'$ represents
values at the conclusion of the change. Hard constraints are
represented by $\inv(v, \xs)$, which may be thought of as
describing invariants of the different control modes. Hybrid automata
are nondeterministic in general; all dynamics compatible with the
automaton's predicates are possible.

Let $A = (V, E, \xs, \init, \flow, \jump, \inv)$ be a hybrid
automaton. We now construct a GSEM $M$ corresponding to $A$. The
endogenous variables are as follows. For each $\xi \in \xs$, $M$ has
real-valued variables $\{X_i^t \mid t \geq 0\}$ corresponding to the
value of $\xi$ at time $t$.  $M$ also has $V$-valued variables $\{S_t
\mid t \geq 0\}$ corresponding to the control mode of the system at
time $t$.
$M$ has a single exogenous variable with a single value, so there is
only one (trivial) context.

The outcomes for intervention $\YY \gets \yy$ are,
similar to ODEs, all assignments to the variables that (1), agree with $\YY \gets \yy$, and (2), are otherwise compatible with the predicates of the automaton. More precisely, $\vv \in M(\uu, \YY \gets \yy)$ iff all the following conditions hold. For convenience, we define functions $\xi(t) = \vv[X_i^t]$ for $i = 1, \dots, n$, $\xs(t) = \left( \dvarv{X}_1(t), \dots, \dvarv{X}_n(t) \right)$, and $\dd(t) = \vv[S_t]$.
\begin{description}
\item[HA1.]
  $\vv[\YY] = \yy$.
\item[HA2.]
$\init(\vv[S_t], \xs(0))$ holds.
\item[HA3.]
  For all $t$, $\inv(\vv[S_0], \xs(t))$ holds.
\item[HA4.]
  For all $t \geq 0$, if $\xs(t)$ is not continuous, at least one of
    (a) or (c) below holds; and if $\dd(t)$ is not continuous, at least one of
(b) or (c) below holds.
  \begin{enumerate}
  \item[(a)] $X_i^t \in \YY$ for some $i = 1, \dots, n$.
  \item[(b)] $S_t \in \YY$.
  \item[(c)] There is an edge $e = (u, v, n) \in E$ such that $\jump(e,
  \lim_{s \to t^-} \xs(t), \xs(t))$ holds.
  \end{enumerate}
\item[HA5.]
Defining \emph{intervention-free} intervals in the same way as in Section \ref{sec:dyn-sys}, the following holds.
For all intervention-free intervals $(a, b)$ such that $\xs$ is continuous on $(a, b)$, we have that $\xs$ is right-continuous at $a$, differentiable on $(a, b)$, and its derivative $\dxs$ satisfies the flow condition $\flow(\vv[S_t], \xs(t), \dxs(t))$ for all $t \in (a, b)$.
\end{description}
Note that HA5 is analogous to the condition ODE3 in Section \ref{sec:dyn-sys}.%
\footnotemark\footnotetext{
  It is possible (and probably desirable) to strengthen HA5 analogously to the way we strengthened ODE3 to ODE3$'$, so that interval interventions on one dynamical variable do not interfere with the flow conditions on another dynamical variable. We do not do this here, because it is not necessary for our simple example (which has only a single variable), and because doing this requires some knowledge of the structure of the predicate $\flow(v, \xs, \dxs)$.
}
The modeler is free to select a set of allowed interventions
that fits the task at hand. In the example below, we choose $\I$ to be
the set of finite compositions of point and interval interventions on
the dynamical variables and control mode variables, but the
outcomes
of $M$ are well-defined for arbitrary interventions. This concludes
the construction of $M$.

A simple thermostat and heater automaton given by
\cite{henzinger_theory_2000} (with the flow conditions slightly
simplified) is as follows. There is a single continuous variable $\temp \in \xs$
and two control modes $V = \{\OFF, \ON\}$. The system starts in OFF with $\temp =
20$, which is the desired set point.
This defines $\init(v, \xs)$.
In OFF, the temperature drifts slowly downward; we have $\dtemp = -0.1$. An invariant of OFF is that
$\temp \geq 18$; if at any point $\temp < 18$, the system transitions to
ON.
Likewise, in ON, the temperature increases rapidly as
$\dtemp = 0.5$,
and an invariant of ON is $\temp \leq 22$.
These conditions define $\flow(v, \xs, \dxs)$ and $\inv(v, \xs)$.
There are two control switches, $e_{\rm on}$ from OFF to ON, and $e_{\rm off}$
from ON to OFF.
Finally, the heater can transition from OFF to ON when $\temp < 19$,
and from ON to OFF when $\temp > 21$. Neither of these transitions
affects the instantaneous value of $\temp$; that is, $\temp' =
\temp$. These conditions define $\jump(e, \xs, \xs')$.

Let us analyze the GSEM $M$ corresponding to this automaton.
In our case, since neither of the two possible discrete jumps change
the value of $\temp$, HA4 implies that $\temp$ must be continuous
everywhere it is not intervened on. Interventions on $\temp$ are
finite
compositions of point and interval interventions.
Furthermore, when $\temp$ is not intervened on, it
cannot change very quickly; by HA5, it must obey the flow condition
(either $\dtemp = -0.1$ or $\dtemp = 0.5$). Hence, given a time
horizon $\tau$, $\temp$ can cross the vertical lines $\temp=19$ and
$\temp=21$ only finitely many times prior to $\tau$. The state of the
heater is specified by the control mode $S^t$. By HA4, the heater
switches ON only at times $t$ when the state of the heater (i.e.,
$S^t$) is intervened on, or when $\temp < 19$; likewise, the heater
switches OFF only when $S^t$ is intervened on, or when $\temp >
21$. Again, interventions on the state of the heater are finite
compositions of point and interval interventions. It follows that the
state of the heater changes only a finite number of times before any
given time horizon $\tau$.
Hence, it is meaningful to talk about the heater discretely
changing state---before
any given
time $\tau$, the heater turns on at $t_1, t_3$
and so on, and turns off at $t_2, t_4$ and so on.

Now that we have some intuition for the behavior of $M$, we examine how $M$ can be used to answer questions of actual cause. By HA2, the heater is initially OFF, and the temperature is initially $\temp = 20$. In the absence of intervention or discrete jumps, the heater will stay OFF and the temperature will drop at the rate of 0.1 per unit time.

Consider an outcome $\vv$ where
the heater does not turn on until, at $t = \frac{18 - 20}{-0.1} = 20$, it is required to do so by HA3; specifically, by the invariant of OFF that
$\temp \geq 18$.
If the heater had been on over any open subinterval $(a, b)$ of $[0, 20)$, the temperature would have been higher than 18 by $t = 20$ by at least $0.5 (b - a)$.
Hence, intuitively, the heater being off over any such subinterval should be considered a cause of $T^{20} = 18$.
However, if we fix any subinterval $(a, b) \subset [0, 20)$ and ask
the formal question of whether
$S(a, b) = \OFF$ is an
actual cause of $T^{20} = 18$ in $(M, \uu, \vv)$ (where $S(a, b) = \{S^t \mid t \in (a, b)\}$),
we run into
problems.%
  \footnote{
  Similar to Footnote \ref{footnote:ac}, there is a technical issue
  here, because the event
  $S(a, b) = \OFF$ (which is an infinite conjunction of equalities) is not in the language $\L(\S)$,
even though the intervention $S(a, b) \gets \OFF$
is.
However, again, we do not view this as a problem, since the
definition of actual causality makes perfect sense for
this formula.
}

AC1 and AC2 both hold, but AC3 does not. AC1 holds, because $\vv[S^t]
= \OFF$ for all $t \in [0, 20)$,
and $\vv[T^{20}] = 18$. AC2 holds, since if we choose $\WW =
  \emptyset$ and
  $\xx' = \ON$
  (recall that $\XX =
  S(a, b)$
  we find that the
  outcome $\vv'$ of $M$ under intervention $\XX \gets \xx'$
  where the heater is on only during $(a, b)$ has $\vv'[T^{20}] = 18 +
  0.5 (b - a) \neq 18$. However, AC3 does not hold, because the open
  subinterval $(a, b)$ contains other open subintervals for which AC1
  and AC2 also hold, by the same arguments.
  This implies that there is
  no
  open interval on which the heater being off is an actual cause of $T^{20} = 18$.

  This creates a dilemma. Since turning the heater on results in $T^{20} > 18$, a good definition of actual cause should provide for some cause.
  In this case, the resolution is that the equality $S^t = \OFF$ for any \emph{point} $t \in (0, 20)$ is in fact an actual cause of $T^{20} = 18$. This is because one of the solutions to $S^t \gets 1$ has $H_s = \ON$ for all $s$ in some nonempty interval starting at $t$, which as before implies $T^{20} > 18$. So AC2 holds. AC1 clearly holds, and AC3 holds because $S^t \gets 1$ is a point intervention, therefore minimal. We believe this resolution to the dilemma is always possible in hybrid automata (if point interventions are allowed), since
    we see no way of defining a hybrid automaton such that when the control mode is intervened on at a point in time, the control mode does not remain at
  the intervened value for some nonzero amount of time in some solution (although we have not attempted to prove this).

  However, we see no reason for this resolution to work in
  general. Other models of dynamical systems may not respond in the
  same way to intervention. This resolution even fails for our example
  hybrid automaton, if it is modified so that point interventions are
  not allowed. In these cases, the definition of actual causality
  presented in Section \ref{sec:ac} fails to provide for any cause of
  $T^{20} > 18$ involving only the heater state. The issue is that AC3
  requires a minimal cause; but minimal causes do not exist in general
  when causes can involve infinitely many variables. It is an open
  problem to find a new definition of actual cause that handles
  infinitely many variables well in general. One potential solution is
  to broaden the set of things that count as causes in infinitary
  settings. In the definition
  presented in Section \ref{sec:ac}, only conjunctions of equalities
  $X = x$ can be actual causes. One could consider expanding possible
  causes to include infinite disjunctions over these equalities, for
  example, the statement that there is \emph{some} nonempty interval
  on which the heater is off:
  \[\exists (a, b) \forall t \in (a, b) S^t = \OFF.\]
  However, we do not pursue this approach further in this paper.

  Note that if we ask instead whether $S^t(a, b) = \OFF$ is an actual cause of $T^{20} \leq 19$ instead, there are no problems. The answer is yes, iff $b - a = 2$. This agrees with the natural
intuition that the heater being off for a sufficiently short time is
not enough to cause the temperature to be low. (There is nothing special about $19$; the solution for any value $x > 18$ is similar.)
\subsection{Rule-based models}

A rule-based model is a dynamical system that transitions
probabilistically between states, with the transition defined by
rewrite rules.
In this
section, we construct a GSEM corresponding to the generic rule-based
model given by Laurent et al.  \nciteyear{laurent_counterfactual_2018}.



Laurent et al. \nciteyear{laurent_counterfactual_2018} show how a
rule-based model can describe a
reaction between a set $S$ of \emph{substrates}
and a set $K$ of \emph{kinases}. The state of the mixture at time $t$
is a binary relation $\bound \subseteq S \times K$ that specifies
which substrates are bound to which kinases, and a unary relation
$\phos \subseteq S \cup K$ that defines which substrates and kinases
have a phosphate group attached.
Chemical interactions between groups of molecules are intended to take place spontaneously, in an analogous fashion to radioactive decay. For example, if at time $t$ there is a substrate $s \in S$ and a kinase $k \in K$ such that $(s, k) \in \bound$ and $s \notin \phos$, then at time $t + \Delta t$, where $\Delta t$ is drawn from an exponential distribution with a time constant that depends on the rule being applied, $s$ will gain a phosphate group (unless in the meantime some other rule has changed the state of the mixture so that the precondition for $s$ and $k$ no longer holds).
These updates are called \emph{events}.
Interventions correspond to blocking some
interactions
from taking place at specific times, for example, ``between
$t=1$ and $t=2$, even if $s$ and $k$ satisfy the above conditions, $s$
cannot gain a phosphate group."



Laurent et al.~explain how to
simulate these dynamics using the following algorithm. For every
possible target $\targ$ of a rule $r$---in the example above,
this would be every substrate-kinase pair $(s, k)$---sample from a
Poisson process with parameter $\tau / \ln(2)$ to obtain a schedule of
times when the rule applies to this target. Then, starting with
the initial mixture and moving through time, whenever any rule
applies according to the schedule, check if the rule's
condition---in our example $(s, k) \in \bound \wedge s \notin
	\phos$---is satisfied for the target, and that the rule is not currently blocked by an intervention.
If these conditions hold, update the mixture
using the rule's mapping (e.g. $\phos \mapsto \phos \cup \{s\}$);
otherwise, do nothing.



This algorithm can immediately be described in a GSEM model. In the
example above, for each time $t \in [0, \infty)$ we would have binary
endogenous variables $\bound^t_{(s, k)}$ for each $s \in S, k \in
	K$, along with variables $\phos_x^t$ for each $x \in S \cup K$. The
exogenous variables correspond to the firing schedule; we have
timestamped variables $T^t_{r, \targ}$, one for each rule $r$, each
target $\targ$ compatible with that rule.
(There are also exogenous variables describing the initial state of the mixture.)
In order to match the
intervention model of \cite{laurent_counterfactual_2018}, we add
additional binary variables $B^t_{r, \targ}$. Intuitively, $B_{r,\targ}^t = 1$ means that the firing of rule $r$ applied to target
$\targ$ at time $t$ is blocked.  Finally, for bookkeeping, we have
binary endogenous variables of the form $X^t_{r, \targ}$
that model whether rule $r$ actually fired on target $\targ$ at time
$t$. The unique outcome is specified in the obvious way:
$X^t_{r, \targ}$ is true
exactly if, at time $t$, $\targ$ satisfies the
condition of $r$,
$T^t_{r, \targ}$ is true, and $B^t_{r, \targ}$ is false. If $X^t_{r, \targ}$
is true, then at time $t$ the state (i.e., the relations
$\bound$ and $\phos$) gets updated using the rule's
mapping. Interventions such as the one above can
simply be described by setting some of the $B^t_{r, \targ}$ to false;
we take the set of allowed interventions $\I$ to be all
interventions of this form. The \emph{trace} $T(\uu, I)$ is simply
the (countable) sequence of variables $X^t_{r, \targ}$ (in ascending
order of $t$) for which $X^t_{r, \targ} = 1$.


Laurent et al. \nciteyear{laurent_counterfactual_2018} defined notions
of \emph{enablement} and
\emph{prevention}.
Enablement and prevention happen at the level of \emph{events}, or
updates to the mixture. Every event $e$ corresponds to a variable
$X^t_{r, \targ}$; it \emph{occurs} if $X^t_{r, \targ} = 1$.  We can
think of the relations $\bound$ and $\phos$ as binary vectors; each
entry in these vectors is called a \emph{site}. For any given event to
occur, certain sites must have certain values. Hence, intuitively,
given two events $e, e'$, $e$ enables $e'$ if $e$ is the last event
before $e'$ that modifies some site to the value that is needed for
$e$ to occur. Likewise, $e$ prevents $e'$ (roughly) if $e$ is the last
event before $e'$ to set a site $s$, and it sets $s$ to a value such
that $e$ cannot occur.
Given a context $\uu$ and an intervention
$I \in \I$, they considered the difference between the trace
$T(\uu)$ and the trace $T(\uu, I)$. They showed that for every element
of the first sequence absent from the intervened sequence, a chain of
enablements and preventions could be traced back from that element to
an element that was directly blocked by $I$. That is, enablements and
preventions were sufficient to explain why each element of $T(\uu)$ no
longer in $T(\uu, I)$ was missing.

The notion of actual cause complements this analysis. For example, it
follows from the sufficiency of enablements and preventions just discussed
that if one rule firing is the actual cause of
another rule firing, then a chain of enablements and preventions can
be traced back from the trace entry for the second rule to the trace
entry for the first. More precisely, if  $B^t_{r, \targ} = 0$ is an
actual cause of $X^{t'}_{r', \targ'} = 1$ in context $\uu$, then a
chain of enablements and preventions can be traced back from
$X^{t'}_{r', \targ'}$ to $X^t_{r, \targ}$ in the pair of traces
$T(\uu)$, $T(\uu, B^t_{r, \targ} \gets 1)$. Without going into the
formalism of \cite{laurent_counterfactual_2018}, a sketch of the proof
of this claim is as follows.  The actual cause statement implies that
$X^{t'}_{r', \targ'}$ is in $T(\uu)$ but not in $T(\uu, B^t_{r, \targ}
\gets 1)$, because the intervention $B^t_{r, \targ} \gets 1$ is the
only one that can satisfy AC2 and AC3. (The other blocking variables
take value zero in both
outcomes
$M(\uu, \emptyset)$ and $M(\uu,
B^t_{r, \targ} \gets 1)$, so setting them to 0 is redundant and
violates AC3.) The only element blocked by $B^t_{r, \targ} \gets 1$ is
$X^t_{r, \targ}$. Hence, a chain of enablements and preventions can be
traced back from $X^{t'}_{r', \targ'}$ to $X^t_{r, \targ'}$.

Actual cause and the GSEM machinery can also be used to answer questions not addressed
by the analysis of Laurent et
al. \nciteyear{laurent_counterfactual_2018}. We
can ask counterfactual questions like ``What would the state at $t=7$
be if every kinase gained a phosphate group at time $t=5$?''
(potentially corresponding to the addition of a test tube's worth of
phosphate solution) or ``Is the fact that substrate $s$ was bound to
kinase $k$ at $t=1$ the actual cause of kinase $k$ gaining a phosphate group
at $t=2$?''
For this reason, we believe that GSEMs are a useful addition to the
rule-based causal modeling toolkit developed
by Laurent et al.
}
\fullv{\sectionDescribingOtherDynamicalSystems}
\section{Conclusion}
\label{sec:conclusion}

While SEMs are a popular modeling framework in
many application areas, they have a restrictive form that makes
working with infinitely many variables difficult. This has led to
attempts to construct application-specific causal models in the study
of ordinary differential equations \cite{BBM19} and
molecular biology
\cite{laurent_counterfactual_2018}.
GSEMs can capture all these applications, while
retaining the input-output behavior of SEMs.
Indeed, SEMs are equivalent to finite GSEMs that satisfy certain properties of SEMs (Halpern's axioms $AX^+$ \cite{Hal20}).
Moreover, GSEMs are easy to use;
converting a
given dynamical model to a GSEM essentially reduces to setting up
allowed interventions, as we demonstrate in examples of ordinary
differential equation models, rule-based models, and hybrid
systems.
Any causal notion defined only in terms of the input-output behavior of SEMs can be straightforwardly carried over to GSEMs. We demonstrate this for the important criterion of \emph{actual cause}, and show how to apply this criterion in examples. Thus, GSEMs are a unifying causal framework that permit modelers to apply the same notions (e.g., actual cause) across many different models of causality.

\shortv{\newpage}
\paragraph{Acknowledgments:}
We thank Sander Beckers for insightful comments on an earlier version of this paper.
Work supported in part by
NSF grants IIS-178108 and IIS-1703846, a grant from the Open
Philanthropy Foundation, ARO grant W911NF-17-1-0592, and MURI grant
W911NF-19-1-0217.

\shortv{\bibliography{joe}}

\appendix



%

\commentout{
	\subsection{Generic SEM for modeling systems of ordinary differential equations}
	\label{dyn-sys-SEM}
	For simplicity, we restrict our attention to a single differential
	equation involving a single variable, but the idea is easily extended
	to generic systems of differential equations as in Section
	\ref{section:dyn-sys}. As before, we assume
	for simplicity all initial-value problems involving this equation have
	unique solutions.
	$$\ddvarv{X} = F(\dvarv{X}).$$
	The dynamics of this equation can be modeled in the following SEM-like model (a SEM where the restrictions of finite $\V$ and finite $\R(X)$ have been dropped, and the structural equations are allowed to be partial, for example by adding a default value).
	Given timestamped variables $X_t, t \in [0, T]$, define partial structural equations $\F_{X_a}(\dots)$ as follows.
	$\F_{X_a}(\{X_b \mid b < a\})$ is the value $v$ for which (1) there exists a time $s < a$ such that
	the values $X_t, s < t < a$, when interpreted as a function
	$\dvarv{X}(t)$, form a solution to the initial-value problem
	$\ddvarv{X} = F(\dvarv{X})$, $\dvarv{X}(0) = X_s$ on the interval $(s,
		t)$, and (2), $\lim_{t \to a-} \dvarv{X}(t) = v$, if such a value
	exists. If no such value exists, $\F_{X_a}(\{X_b \mid b < a\})$ is
	undefined. Note that $\F_{X_a}$ is well defined since given this
	definition, if there is such a value $v$, it is necessarily unique.
}

\fullv{
\clearpage

\section{An axiom system for causal reasoning}
\label{section:axioms}

We now review the axiom systems considered by Halpern \nciteyear{Hal20}
for reasoning about causality.
Note that there are two slight differences between our presentation
and that of Halpern.
First, as we mentioned earlier, we have weakened the
language of causal formulas so that
primitive
causal formulas are no
longer parameterized by
contexts. Thus, our language has formulas such as $[\YY \gets \yy](X
	= x)$ rather than $[\YY \gets \yy](X(\uu) = x)$.
Second, the list of axioms given below does not include two of
Halpern's axioms, which he called D10 and D11.
D11 is a technical axiom that was needed only to reason
about formulas with contexts (to reduce to formulas that mentioned ony
one context); D10 says that there are unique
outcomes, and is
redundant in acyclic systems.
A minor modification of Halpern's proof shows that
the axiom systems $AX^+$ and  $AX^+_{rec}$ defined below (which are
identical to the
system Halpern called $AX^+$ and $AX^+_{rec}$, respectively, except
that they omit the axioms D10 and D11)
are sound and complete for SEMs and acyclic SEMs, respectively, with
respect to the language that we are considering
(just as Halpern's versions of $AX^+$ and $AX^+_{rec}$ were sound and
complete for his language); the proof is essentially identical to
Halpern's, so we omit it here.
To axiomatize acyclic SEMs, following Halpern, we define $Y
	\leadsto Z$, read ``$Y$ affects $Z$'', as an abbreviation for the
formula
\begin{equation} \nonumber
	\begin{split}
		&\lor_{\XX \subseteq \V, \xx \in \R(\XX), y \in \R(y), z \ne z' \in \R(Z)} \\
		&\quad ([\XX \gets \xx](Z = z) \land [\XX \gets \xx, Y \gets y](Z = z'));
	\end{split}
\end{equation}
that is, $Y$ affects $Z$
if there is some setting of some endogenous variables $\XX$
for which changing the value of $Y$ changes the
value of $Z$.  This definition is used in axiom D6 below, which
characterizes
acyclicity.

\begin{definition}
	$AX^+$ consists of axiom schema D0-D5 and D7-D9, and inference rule MP.
	$AX^+_{rec}$ results from adding D6 to $AX^+$.
\end{definition}

\begin{itemize}
	\item[D0.] All instances of propositional tautologies.
	\item[D1.] $[\YY \gets \yy](X = x \rimp
		      X \ne x')$  if $x, x' \in \R(X)$, $x \ne x'$ \hfill
	      (functionality)
	\item[D2.] $[\YY \gets \yy](x \in \R(X))$
	      \hfill (definiteness)
	\item[D3.] $\<\XX \gets \xx\>(W = w
		      \land \YY = \yy)
		      \rimp \<\XX \gets \xx;W \gets
		      w\>(\YY = \yy)$
	      \hfill
	      (composition)
	\item[D4.] $[\WW \gets \boldw; X \gets x](X = x)$ \hfill
	      (effectiveness)
	\item[D5.] $(\<\XX \gets \xx; Y \gets y\> (W = w \land
		      \ZZ = \zz)  \land
		      \<\XX \gets \xx; W \gets w\> (Y = y \land
		      \ZZ = \zz))$\\
	      $\mbox{ }\ \ \ \rimp \<\XX \gets \xx\> (W
		      = w \land Y = y \land \ZZ = \zz)
	              $, where $\ZZ = \V - (\XX \union \{W,Y\})$
	              \mbox{ } \hfill
	      (reversibility)
	\item[D6.]
	      $(X_0 \leadsto X_1 \land \ldots \land X_{k-1} \leadsto X_k)
		      \rimp
		      \neg (X_k \leadsto X_0)$
	      \hfill
	      (recursiveness)
	\item[D7.] $([\XX \gets \xx]\phi \land [\XX \gets
			      \xx](\phi \rimp \psi)) \rimp  [\XX \gets \xx]\psi$
	      \hfill
	      (distribution)

	\item[D8.] $[\XX \gets \xx]\phi$ if $\phi$ is a propositional
	      tautology  \hfill (generalization)
	\item[D9.]
	      $\<\YY \gets \yy\>true \wedge \big( \<\YY \gets \yy\>(X = x) \Rightarrow \<Y \gets y\>(X \neq x') \big)$,
	      if $x \neq x'$.
        \mbox{ } \hfill
(unique outcomes for $\V - \{X\}$)
	\item[MP.] From $\phi$ and $\phi \rimp \psi$, infer $\psi$
	      \hfill
	      (modus ponens)
\end{itemize}


\section{Proofs}
\label{appendix:additional-proofs}

\newenvironment{oldthm}[1]{\par\noindent{\bf Theorem #1:} \em \noindent}{\par}
\newenvironment{oldlem}[1]{\par\noindent{\bf Lemma #1:}
	\em \noindent}{\par}
\newenvironment{oldcor}[1]{\par\noindent{\bf Corollary #1:} \em \noindent}{\par}
\newenvironment{oldpro}[1]{\par\noindent{\bf Proposition #1:} \em \noindent}{\par}
\newcommand{\othm}[1]{\begin{oldthm}{\ref{#1}}}
		\newcommand{\eothm}{\end{oldthm} \medskip}
\newcommand{\olem}[1]{\begin{oldlem}{\ref{#1}}}
		\newcommand{\eolem}{\end{oldlem} \medskip}
\newcommand{\ocor}[1]{\begin{oldcor}{\ref{#1}}}
		\newcommand{\eocor}{\end{oldcor} \medskip}
\newcommand{\opro}[1]{\begin{oldpro}{\ref{#1}}}
		\newcommand{\eopro}{\end{oldpro} \medskip}

\othm{theorem:satisfies-same-formulas-equiv-has-same-solutions}
If $M$ and $M'$ are causal models over the same signature
$\S$ that,  given a context and intervention, return a finite set
of outcomes, then $M$ and $M'$ have the same outcomes (that
is, for all $\uu \in \R(\U)$ and $I \in \I$, $M(\uu, I) =
	M'(\uu, I)$) if and only if they satisfy the same set of causal
formulas (that is, for all $\uu \in \R(\U)$ and $\psi \in \L(\S)$,
$M, \uu \models \psi \Leftrightarrow M', \uu \models \psi$).
\eothm
\prf
Let $M$ and $M'$ be causal models with the same set of solutions. It
suffices to consider the
primitive
causal formulas \mbox{$[\YY \gets \yy](X =
		x)$}, since the truth of other formulas in $\L(\S)$ are derived from
  these. Recall that $M, \uu \models{[\YY \gets \yy](X = x)}$ iff for
all outcomes $\vv \in M(\uu, \YY \gets \yy)$, $\vv[X] = x$. But
$M(\uu, \YY \gets \yy) = M'(\uu, \YY \gets \yy)$, so $M, \uu
\models [\YY \gets \yy](X = x)$ if and only if
$M', \uu \models [\YY	\gets \yy](X = x)$.
Conversely, suppose that $M$ and $M'$ satisfy
the same set of causal formulas. Suppose for contradiction that there
exists some $\uu$ and $I$ with $M(\uu, I) \neq M'(\uu, I)$. Then
without loss of generality, there is an outcome $\vv$ in $M(\uu, I)$
that is not in
$M'(\uu, I)$.
This outcome must differ from each of the finitely many outcomes
$\{\vv_1, \vv_2, \dots, \vv_n\} = M'(\uu, I)$ in at
least one variable; that is, there must be variables $X_1, X_2, \dots,
	X_n \in \V$ with $\vv[X_i] \neq \vv_i[X_i]$ for $i \in 1, 2, \dots,
	n$. Consider the causal formula $\varphi = \<I\>(\band_{1 \leq i \leq
	  n} X_i = \vv[X_i])$. We have that $M, \uu \models
          \varphi$, since
$\vv \in M(\uu, I)$. However,
it is not true that $M', \uu \models \varphi$,
because no outcome $\vv_i$ of $M'$ satisfies
$\band_{1 \leq i \leq  n} X_i = \vv[X_i]$. This contradicts the assumption
s          that $M$ and $M'$ satisfy
the same set of causal formulas; hence $M(\uu, I) = M'(\uu, I)$ for
all $\uu \in \R(U)$, $I \in \I$.
\eprf

\othm{thm:GSEM-generalizes-SEM}
For all SEMs $M$, there is a GSEM $M'$ such that $M \equiv M'$.
\eothm
\prf
Given a SEM $M = ((\U, \V, \R, \I), \F)$, define
$M' = ((\U, \V, \R, \I), \F')$, where
for all $\XX \gets \xx \in
	\I$ and $\uu \in \U$,
$\F'(\uu, \XX \gets \xx) =
	M(\uu, \XX \gets \xx)$.
Since $M'(\uu, \XX \gets \xx) = \F'(\uu, \XX \gets \xx)$, $M'$ is equivalent to $M$ by definition.
\eprf

\commentout{
begin{theorem}
label{theorem:complete-interventions-and-axioms-decide-everything}
If $\V$ is finite and $\R(X)$ is finite for each $X \in \V$, the
axiom system $AX$, along with exactly one formula of the form $[(\V
			\setminus X) \gets \yy](X = x(\yy))$ for every variable $X$ and
setting $\yy \in \R(\V \setminus X)$ forms a consistent set of
premises from which the truth and falsity of every other causal
formula is provable.
\end{theorem}
Note: the notation $x(\yy)$ is intended to show the dependence of the unique value $x \in \R(X)$ on the specific intervention $\YY \gets \yy$.

\prf
This is a reformulation of (1) the completeness of $AX$ for SEMs, and (2) the fact that SEMs are defined in terms of complete interventions. The formulas specified above give the value of each variable $X$ under every complete intervention (on $\V \setminus X$). These formulas can be used to define a SEM $M$ as follows. Take $M$ to have no exogenous variables, (or more precisely, that there is only one context $\uu$, so that there is no distinction between $M \models \varphi$ and $M, \uu \models \varphi$), and define $\F_X(\yy) = x(\yy)$. Notice that $M$ models the causal formula $\psi$ defined by
$$\psi = \wedge_{X \in \V} \wedge_{\yy \in \R(\YY)} [\YY \gets \yy](X = x(\yy)),$$
%
where $\YY = \V - X$ and $x(\yy)$ denotes the unique value $x \in \R(X)$ such that the conjunct is in the set of premises. Note that this formula makes sense because $\V$ is finite (which implies $\YY$ is finite), and $\R(X)$ is finite for each variable $X$ (which implies $\R(\YY)$ is finite).

Thus, for any causal formula $\varphi$, if $M \models \varphi$, $M$
models the causal formula $\psi \Rightarrow \varphi$, and if $M
	\models \neg \varphi$, then $M \models \psi \Rightarrow \neg
	\varphi$. For all causal formulas $\varphi$, one of these two cases
holds, since a
causal formula is either true or false in a given SEM under a given
context.

However---and this is the key insight---$M$ is the only SEM over signature $\S$ (up to equivalence) such that $M \models \psi$. Hence, the implications $\psi \Rightarrow \varphi$ and $\psi \Rightarrow \neg \varphi$ above are valid for all SEMs. Completeness of $AX$ for SEMs over $\L(\S)$ then implies that these implications are provable from $AX$. But the hypothesis of these implications is merely the conjunction of our premises. Hence given the premises, we can derive the conclusion using MP. To recap, for any causal formula $\varphi$, either $\psi \Rightarrow \varphi$ or $\psi \Rightarrow \neg \varphi$ is valid for all SEMs, hence provable from $AX$. Thus using MP, either $\varphi$ or $\neg \varphi$ is provable using both $AX$ and the premises $\psi$.
\eprf
}


\othm{theorem:eqn-modification-equiv-intervention-composition-SEM}
For all SEMs $M$ and interventions $I, J \in \I$ such that $I ; J
	\in \I$, we have that $M_I(\uu, J) = M(\uu, I ; J)$.
\eothm
\prf
We prove the equivalent statement $(M_I)_J(\uu) = M_{I ; J}(\uu)$.
Since the outcomes of SEMs are determined by the structural
equations, it suffices to show that the structural equations of
$(M_I)_J$ are the same as those of $M_{I ; J}$. Let $X \in \V$ be
arbitrary and consider the structural equation $\F_X$. Without loss of
generality, let $I = \YY \gets \yy$ and $J = \ZZ \gets \zz$. There are
three cases to consider: $X \notin \YY \cup \ZZ$, $X \in \ZZ$, and $X
	\in \YY \setminus \ZZ$. The first case is trivial; $\F_X$ is
unmodified in both models. In the second case, letting $\bold{s} \in
	\R(\V \setminus \{X\})$ denote an arbitrary input to $\F_X$, in
$(M_I)_J$, we have that $\F_X(\bold{s}) = \ZZ[X]$. But by the
definition of $I ;
	J$, we also have $\F_X(\bold{s}) = \ZZ[X]$ in $M_{I ; J}$. In the third case,
in $(M_I)_J$, $\F_X(\bold{s}) = \YY[X]$, since in $M_I$,
$\F_X(\bold{s}) = \YY[X]$, and applying the intervention $J$ does not
affect $\F_X$ since $X \notin \ZZ$.
\eprf

\othm{theorem:intervened-models}
  Suppose that $M$ and $M'$ are causal models with $M \equiv M'$.
Then for all $I \in \I$, we have that $M_I \equiv M'_I$.
\eothm
\prf
Clearly $M_I$ and $M'_I$ have the same signatures. It remains to
show that for all contexts $\uu$ and all intervention $J$ allowed
in $M_I$, we have that $M_I(\uu, J) = M'_I(\uu, J)$.
Applying the definition and the fact that $M \equiv M'$, we have that $M'_I(\uu, J) = M'(\uu, I ; J) = M(\uu, I ; J)$. Therefore, it suffices to show $M(\uu, I ; J) = M_I(\uu, J)$, which is exactly Theorem \ref{theorem:eqn-modification-equiv-intervention-composition-SEM}.
\eprf

The following theorem is needed to prove Theorem
\ref{theorem:SEMs-are-equivalent-to-finite-GSEMs-if-all-interventions-allowed}.
\begin{theorem} \label{theorem:equivalence-of-finite-models-satisfying-AX-and-agreeing-on-complete0interventions}
	If $M$ and $M'$ are causal models (either SEMs or GSEMs) with a common signature
	$\S=(\U,\V,\R,\I_{univ})$,
	where $\V$ is finite and $\R(X)$ is finite for all $X
		\in \V$, that both satisfy the axioms in $AX^+$ and have the
same outcomes under complete interventions---that is, for all
	$\uu \in \R(\U)$ and $X \in \V$, if $\YY = \V \setminus X$, then
	for all $\yy \in \R(\YY)$,
	$M(\uu, \YY \gets \yy) = M'(\uu, \YY \gets \yy)$---%

	then $M$ and $M'$ agree on all causal formulas.
\end{theorem}
\prf
Fix an arbitrary context $\uu$. $M$ satisfies axiom D9, so for every variable $X \in \V$,
and for every assignment $\yy \in \R(\YY)$ to the variables $\YY = \V \setminus \{X\}$,
there is a unique $x \in \R(X)$ such that $M, \uu \models [\YY \gets \yy](X = x)$.
Using this fact, we can define a SEM $M''$ with signature $\S$ as follows.
Define $\F''_X(\uu, \yy)$ to be the unique $x$ such that
$M, \uu \models [\V \setminus \{X\} \gets \yy](X = x)$.
Let $C$ be the set of all formulas $\varphi = [\V \setminus \{X\} \gets \yy](X = x)$ such that $M, \uu \models \varphi$.
By assumption, $C$ is also the set of all such formulas $\varphi$ for which $M', \uu \models \varphi$.
Let $\chi$ be the conjunction of all the formulas in $C$. Since there are finitely many variables, and all ranges are finite, this set of formulas is finite, and so taking the conjunction makes sense.
We know that $M$ and $M'$ satisfy all axioms of $AX^+$, and both
models satisfy $\chi$. This means that if $\chi \rimp \psi$ is provable in
$AX^+$, then $M$ and $M'$ both satisfy $\psi$.
We now show that,
for all formulas $\psi$, either $\chi \rimp \psi$ or $\chi \rimp \neg
	\psi$ is provable
in $AX^+$. This means that either both $M, \uu \models
	\psi$ and $M', \uu \models \psi$ (if $\chi \rimp \psi$ is provable in
$AX^+$), or both $M, \uu \nvDash \psi$ and $M, \uu
	\nvDash \psi$ (if $\chi \rimp \neg \psi$ is provable in
$AX^+$). That is, $M$ and $M'$ agree on all causal formulas.
Note that $\chi$ is false in all SEMs over $\S$ other than models that
agree with the $M''$ that we defined using $\chi$ in context $\uu$.
Thus, if $(M'',\uu) \models \psi$, then $\chi \rimp \psi$ is valid;
and if $(M'',\uu) \models \neg \psi$, then $\chi \rimp \neg \psi$ is valid.
Since $AX^+$ is a sound and complete axiomatization, it follows that
either $\chi \rimp \psi$ or $\chi \rimp \neg \psi$ is provable, as desired.
\eprf

\othm{theorem:SEMs-are-equivalent-to-finite-GSEMs-if-all-interventions-allowed}
For all finite GSEMs  over a signature $\S$ such that
$\I = \I_{univ}$ and all the axioms of $AX^+$ are
        valid, there is an equivalent SEM,
and vice versa.
\eothm
\prf
Given a SEM $M$, define a
GSEM $M'$ with the same signature by taking $\FF'(\uu, I) = M(\uu, I)$, as
in Theorem \ref{thm:GSEM-generalizes-SEM}. This GSEM is clearly
equivalent to $M$. Furthermore, all the axioms in $AX^+$ are valid in $M$.
This follows from the facts that (1) equivalent causal models have the
same outcomes (by definition),
(2) finite causal models with the same
outcomes satisfy the same causal formulas (Theorem
\ref{theorem:satisfies-same-formulas-equiv-has-same-solutions}), and
(3) $M$ is a SEM, so all the axioms in $AX^+$ are valid in $M$.
Conversely, given a finite GSEM $M'$ in which all the axioms of $AX^+$
are valid, the GSEM must
have unique solutions for $\V \setminus X$ (D9). That is, for each
context $\uu \in \R(U)$ and each variable $X \in \V$, if we define
$\YY = \V \setminus X$, for every $\yy \in \R(Y)$,
there is a unique $x \in \R(X)$ such that $M', \uu \models [\YY \gets
		\yy](X = x)$.
Here we use the fact that $\I = \I_{univ}$ to ensure that the relevant
instances of
D9 are in the language.
We can use this property to define the structural equations of the SEM $M$. That is, define a SEM $M$ with the same signature by defining $\F_X(\uu, \yy) = x$, where $x$ is the value guaranteed above.
We must show that $M$ has the same outcomes as $M'$. But this is
just Theorem
\ref{theorem:equivalence-of-finite-models-satisfying-AX-and-agreeing-on-complete0interventions}.
\eprf

\othm{theorem:GSEMs-and-acyclic-SEMs}
For all finite GSEMs over a signature $\S$
such that $\I = \I_{univ}$ and all the axioms $AX^+_{rec}$ are
valid, there is an equivalent acyclic SEM, and
vice versa.
\eothm
\prf
Given a finite GSEM $M'$ satisfying
$AX^+_{rec}$, Theorem
\ref{theorem:SEMs-are-equivalent-to-finite-GSEMs-if-all-interventions-allowed}
guarantees the existence of an equivalent SEM $M$.
Since $M$ is equivalent to $M'$, $M$ satisfies $AX^+_{rec}$. This
implies that $M$ is acyclic. To prove this, suppose not. Then there is
$k > 1$ and endogenous variables $V_1, \dots, V_k$ having cyclic
dependencies; that is, $V_{i + 1}$ is not independent of $V_i$ for $i
= 1, \dots, k-1$, and $V_1$ is not independent of $V_k$. But it is
easy to see that if $Y$ is not independent of $X$, then $X$ affects
$Y$, i.e., $X \leadsto Y$. Thus, $V_k \leadsto V_1 \wedge V_1 \leadsto
V_2 \wedge \dots \wedge V_{k-1}\leadsto V_k$. This is the negation of
an instance of D6. Hence not all the axioms of $AX^+_{rec}$ are valid
in $M$,  a
contradiction.
For the converse, given an acyclic SEM $M$, Theorem
\ref{thm:GSEM-generalizes-SEM} guarantees the existence of an
equivalent GSEM $M'$. This equivalent GSEM satisfies the same formulas
as $M$, so it satisfies $AX^+_{rec}$.
\eprf

\othm{theorem:solve-ode-gsem}
The set of outcomes output by valid executions of
SOLVE-ODE-GSEM are exactly the outcomes $M_{\ODE}(\uu, I)$.
\eothm
\prf
We walk through the algorithm's execution and show that
whenever it defines a dynamical variable (and thus a model
variable, via the translation in step 4),
it can make all the choices compatible with ODE1, ODE2, and ODE3$'$, and
cannot make any other choices:
%
\begin{itemize}
  \item
In step 1,
$\dvar{i}(0) = X_i^0$ is the only choice consistent with the
right-continuity requirement of ODE3$'$.
%
  \item
In step 2(a), $\dvarv{X}_j(t) = k$ is the only choice consistent with
ODE1. It is compatible with ODE2 and ODE3, since ODE2 and ODE3 require
nothing of intervened points.
%
  \item
In step 2(b) and step 3, the possible settings for the intervention-free variables are exactly the settings allowed by ODE3$'$.
They are compatible with ODE1, since the intervention-free variables
are not intervened on in $(a, b)$, and compatible with ODE2, since
solutions to initial value problems are always continuous.
%
  \item
In step 2(c)(i), $\dvarv{X}_j(t_i) = \xx[X_j^{t_i}]$ is the only choice consistent with ODE1. It is compatible with ODE2 and ODE3 since, again, ODE2 and ODE3 require nothing of intervened points.
%
\item
  In step 2(c)(ii), $\dvar{j}(t_i) = \lim_{t \to t_i^-} \dvar{j}(t)$ is the only choice that maintains left-continuity (is consistent with ODE2). It is compatible with ODE1, since $\dvar{j}$ is not intervened on at time $t_i$, and compatible with ODE3, since by construction, there is no intervention-constant open interval containing $t_i$.
  Finally, the limit always exists, because  the values of $\dvar{j}$ on $(t_{i-1}, t_i)$ were set in step 2(b), so $\dvar{j}$ is continuous on the open interval $(t_{i-1}, t_i)$.
\end{itemize}
\eprf

\othm{theorem:CCMs-equivalent-to-GSEMs}
For all GSEMs, there is an equivalent
CCM, and vice versa.
\eothm
\prf
Given a CCM $M' = (\S, \C, I)$, define a GSEM $M = (\S, \FF, \I)$ by taking
$\FF(\uu, I) = M'(\uu, I)$; it is immediate
that $M$ and $M'$ have the same outcomes.
For the converse, given
a GSEM $M = (\S, \FF, \I)$, define a CCM $M' = (\S, \C, \I)$ as
follows.
For every intervention $\XX \gets \xx \in \I$, $\C$ contains a
constraint $C_{\XX \gets \xx}$ such that $a_{C_{\XX \gets \xx}} = \{\XX\}$, and
for every context $\uu$,
$f_{C_{\XX \gets \xx}}(\uu, \vv) = 1$ iff either $\vv[\XX] \neq \xx$ or $\vv \in
\FF(\uu, \XX \gets \xx)$.
We claim that the outcomes $M'(\uu, \XX \gets \xx)$  are exactly the
GSEM outcomes $\FF(\uu, \XX \gets \xx)$. Indeed, suppose $\vv \in
M'(\uu, \XX \gets \xx)$. Then $\vv[\XX] = \xx$, and it follows from
the constraint $C_{\XX \gets \xx}$ that $\vv \in
\FF(\uu, \XX \gets \xx)$. For the opposite implication, suppose that $\vv \in
\FF(\uu, \XX \gets \xx)$. Then $\vv[\XX] =
\xx$ (since GSEMs satisfy effectiveness). Moreover, $\vv$ satisfies
all active constraints. It satisfies the constraint corresponding to
$\XX \gets \xx$, since $\vv \in \F(\uu, I)$. And it satisfies the
constraints $C_{\XX \gets \xx'}$ for $\xx' \neq \xx$,
since $\vv[\XX] = \xx \neq \xx'$.
\eprf


                \begin{theorem}
	\label{theorem:shell-game-satisfies-AX}
	$M_{shell}$ satisfies all the axioms in $AX^+$.
\end{theorem}
\prf
D0, D1, D2, D7 and D8 are trivial. No joint interventions are
allowed, so the only way to instantiate D3 is to have $\XX = W = S_1$
(or symmetrically, $\XX = W = S_2$). But if $\XX = W$, then $\XX
	\gets \xx; W \gets w = W \gets w$ and D3 follows trivially by
eliminating the conjunction. D4 (effectiveness) holds by
inspection. D5 holds for the same reason as D3. Finally, D9 cannot be
instantiated because no complete interventions are allowed.
\eprf

\shortv{\sectionDescribingOtherDynamicalSystems}

 \bibliographystyle{named}
 \bibliography{joe}
} 

\end{document}

%% file: defn.tex

\newtheorem{THEOREM}{Theorem}[section]
\newenvironment{theorem}{\begin{THEOREM} \hspace{-.85em} {\bf :} }%
                        {\end{THEOREM}}
\newtheorem{LEMMA}[THEOREM]{Lemma}
\newenvironment{lemma}{\begin{LEMMA} \hspace{-.85em} {\bf :} }%
                      {\end{LEMMA}}
\newtheorem{COROLLARY}[THEOREM]{Corollary}
\newenvironment{corollary}{\begin{COROLLARY} \hspace{-.85em} {\bf :} }%
                          {\end{COROLLARY}}
\newtheorem{PROPOSITION}[THEOREM]{Proposition}
\newenvironment{proposition}{\begin{PROPOSITION} \hspace{-.85em} {\bf :} }%
                            {\end{PROPOSITION}}
\newtheorem{DEFINITION}[THEOREM]{Definition}
\newenvironment{definition}{\begin{DEFINITION} \hspace{-.85em} {\bf :} \rm}%
                            {\end{DEFINITION}}
\newtheorem{CLAIM}[THEOREM]{Claim}
\newenvironment{claim}{\begin{CLAIM} \hspace{-.85em} {\bf :} \rm}%
                            {\end{CLAIM}}
\newtheorem{EXAMPLE}[THEOREM]{Example}
\newenvironment{example}{\begin{EXAMPLE} \hspace{-.85em} {\bf :} \rm}%
                            {\end{EXAMPLE}}
\newtheorem{REMARK}[THEOREM]{Remark}
\newenvironment{remark}{\begin{REMARK} \hspace{-.85em} {\bf :} \rm}%
                            {\end{REMARK}}

\newcommand{\thm}{\begin{theorem}}
\newcommand{\lem}{\begin{lemma}}
\newcommand{\pro}{\begin{proposition}}
\newcommand{\dfn}{\begin{definition}}
\newcommand{\rem}{\begin{remark}}
\newcommand{\xam}{\begin{example}}
\newcommand{\cor}{\begin{corollary}}
\newcommand{\prf}{\noindent{\bf Proof:} }
\newcommand{\ethm}{\end{theorem}}
\newcommand{\elem}{\end{lemma}}
\newcommand{\epro}{\end{proposition}}
\newcommand{\edfn}{\bbox\end{definition}}
\newcommand{\erem}{\bbox\end{remark}}
\newcommand{\exam}{\bbox\end{example}}
\newcommand{\ecor}{\end{corollary}}
\newcommand{\eprf}{\bbox\vspace{0.1in}}
\newcommand{\beqn}{\begin{equation}}
\newcommand{\eeqn}{\end{equation}}

\newcommand{\bbox}{\vrule height7pt width4pt depth1pt}

\newcommand{\clm}{\begin{claim}}
\newcommand{\eclm}{\end{claim}}






\newcommand{\boldw}{{\bf w}}




\newcommand{\rimp}{\Rightarrow}



\newcommand{\band}{\bigwedge}
\newcommand{\union}{\cup}

\newcommand{\xx}{{\bf x}}
\newcommand{\yy}{{\bf y}}
\newcommand{\uu}{{\bf u}}
\newcommand{\vv}{{\bf v}}
\newcommand{\FF}{{\bf F}}



\renewcommand{\phi}{\varphi}





\newcommand{\C}{{\cal C}}

\newcommand{\F}{{\cal F}}

\newcommand{\I}{{\cal I}}
\newcommand{\J}{{\cal J}}

\renewcommand{\P}{{\cal P}}

\newcommand{\R}{{\cal R}}

\newcommand{\U}{{\cal U}}
\newcommand{\V}{{\cal V}}

\newcommand{\WW}{{\bf W}}
\newcommand{\XX}{{\bf X}}
\newcommand{\YY}{{\bf Y}}
\newcommand{\ZZ}{{\bf Z}}
\newcommand{\zz}{{\bf z}}
\newcommand{\ww}{{\bf w}}


\newcommand{\<}{\langle}
\renewcommand{\>}{\rangle}

\newcommand{\ol}{\setlength{\itemsep}{0pt}\begin{enumerate}}
\newcommand{\eol}{\end{enumerate}\setlength{\itemsep}{-\parsep}}
\newcommand{\ul}{\setlength{\itemsep}{0pt}\begin{itemize}}
\newcommand{\dl}{\setlength{\itemsep}{0pt}\begin{description}}
\newcommand{\edl}{\end{description}\setlength{\itemsep}{-\parsep}}
\newcommand{\eul}{\end{itemize}\setlength{\itemsep}{-\parsep}}





\setcounter{secnumdepth}{2} 











\newcommand{\commentout}[1]{}

\newcommand{\bi}{\begin{itemize}}
\newcommand{\ei}{\end{itemize}}
\newcommand{\be}{\begin{enumerate}}
\newcommand{\ee}{\end{enumerate}}